\documentclass{article}

\usepackage{microtype}
\usepackage{graphicx}
\usepackage{subfigure}
\usepackage{booktabs} % for professional tables
\usepackage{tabularray}
\usepackage{multicol}
\usepackage{multirow}
\usepackage{xfrac}
\usepackage{nicematrix}
\usepackage[font=small,labelfont=bf]{caption}
\usepackage{xspace}
\usepackage{algorithm}
\usepackage{algpseudocode}
\usepackage{xcolor,cancel}
\usepackage{enumitem}

\newcommand\hcancel[2][black]{\setbox0=\hbox{$#2$}%
\rlap{\raisebox{.45\ht0}{\textcolor{#1}{\rule{\wd0}{1pt}}}}#2} 

\usepackage{hyperref}

\usepackage[accepted]{icml2024}

\usepackage{amsmath,amsfonts,bm,mathtools,microtype,amsthm}

\theoremstyle{plain}
\newtheorem{theorem}{Theorem}[section]
\newtheorem{definition}[theorem]{Definition}

\newtheorem{lemma}[theorem]{Lemma}

\newtheorem{assumption}[theorem]{Assumption}

\def\eqref#1{equation~\ref{#1}}
\def\1{\bm{1}}

\def\vtheta{{\bm{\theta}}}
\def\vpsi{{\bm{\psi}}}
\def\vphi{{\bm{\phi}}}
\def\va{{\bm{a}}}

\def\vq{{\bm{q}}}

\def\vs{{\bm{s}}}

\def\vx{{\bm{x}}}
\def\vy{{\bm{y}}}

\def\vq{{\bm{q}}}

\DeclareMathOperator{\diag}{diag}
\newcommand{\erf}[1]{\operatorname{erf}\left(#1\right)}
\newcommand{\erfc}[1]{\operatorname{erfc}\left(#1\right)}

\DeclareMathAlphabet{\mathsfit}{\encodingdefault}{\sfdefault}{m}{sl}
\SetMathAlphabet{\mathsfit}{bold}{\encodingdefault}{\sfdefault}{bx}{n}

\newcommand{\R}{\mathbb{R}}

\newcommand{\norm}[1]{\left\lVert#1\right\rVert}

\DeclarePairedDelimiter{\defaultDelim}{[}{]}

\DeclareMathOperator{\capE}{\mathbb{E}}
\newcommand{\E}[2][]{\capE_{#1}\defaultDelim*{#2}}
\DeclareMathOperator{\capVar}{Var}
\newcommand{\Var}[2][]{\capVar_{#1}\defaultDelim*{#2}}

\usepackage[capitalize,noabbrev]{cleveref}

\usepackage[textsize=tiny]{todonotes}
\usepackage{soul}

\usepackage{caption}
\newcommand{\paperTitle}{Adaptive Horizon Actor-Critic for Policy Learning in Contact-Rich Differentiable Simulation\xspace}

\icmltitlerunning{\paperTitle}
\begin{document}
\twocolumn[
\icmltitle{Adaptive Horizon Actor-Critic for Policy Learning \\ in Contact-Rich Differentiable Simulation}



\icmlsetsymbol{equal}{*}

\begin{icmlauthorlist}
\icmlauthor{Ignat Georgiev}{gt}
\icmlauthor{Krishnan Srinivasan}{stanford}
\icmlauthor{Jie Xu}{nvidia}
\icmlauthor{Eric Heiden}{nvidia}
\icmlauthor{Animesh Garg}{gt,nvidia,toronto}
\end{icmlauthorlist}

\icmlaffiliation{gt}{Georgia Institute of Technology}
\icmlaffiliation{stanford}{Stanford University}
\icmlaffiliation{nvidia}{Nvidia}
\icmlaffiliation{toronto}{University of Toronto}

\icmlcorrespondingauthor{Ignat Georgiev, Animesh Garg}{\{ignat, animesh.garg\}@gatech.edu}

\icmlkeywords{Machine Learning, ICML}

\vskip 0.3in
]



\printAffiliationsAndNotice{}  % leave blank if no need to mention equal contribution

\begin{abstract}
Model-Free Reinforcement Learning~(MFRL), leveraging the policy gradient theorem, has demonstrated considerable success in continuous control tasks. However, these approaches are plagued by high gradient variance due to zeroth-order gradient estimation, resulting in suboptimal policies. Conversely, First-Order Model-Based Reinforcement Learning~(FO-MBRL) methods employing differentiable simulation provide gradients with reduced variance but are susceptible to sampling error in scenarios involving stiff dynamics, such as physical contact. This paper investigates the source of this error and introduces Adaptive Horizon Actor-Critic (AHAC), an FO-MBRL algorithm that reduces gradient error by adapting the model-based horizon to avoid stiff dynamics. Empirical findings reveal that AHAC outperforms MFRL baselines, attaining 40\% more reward across a set of locomotion tasks and efficiently scaling to high-dimensional control environments with improved wall-clock-time efficiency. \href{https://adaptive-horizon-actor-critic.github.io/}{{adaptive-horizon-actor-critic.github.io}}
\end{abstract}

\section{Introduction}

The Policy Gradients Theorem \citep{sutton1999policy} has enabled the development of Model-Free Reinforcement Learning~(MFRL) approaches for solving continuous motor control tasks. Although these methods have achieved impressive results \citep{hwangbo2017control,akkaya2019solving,hwangbo2019learning}, they are hampered by high gradient variance leading to unstable learning and suboptimal policies \citep{mohamed2020monte}, as well as subpar sample efficiency  \citep{amos2021model}. The latter can be circumvented via the use of efficient vectorized physics simulators. These simulators, when integrated with efficient MFRL methods, facilitate rapid training such as learning quadruped walking in minutes \citep{rudin2022learning}. However, the effectiveness of MFRL in addressing motor control challenges, even with extensive data, remains questionable.

\begin{figure}[t]
  \includegraphics[width=\linewidth]{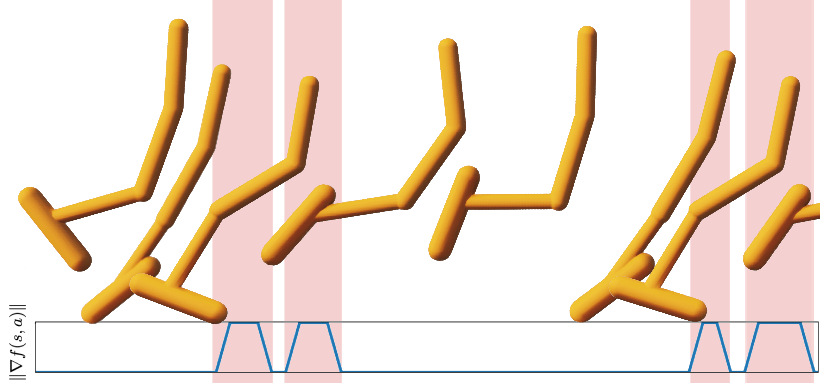}
  \caption{\textbf{Overview.} We find that First Order Model-Based RL (FO-MBRL) methods suffer from erroneous gradients arising from stiff dynamics $\left( \norm{\nabla f(s,a)} \gg 0 \right)$. Our proposed method, \textit{AHAC}, truncates model-based trajectories at the point of contact, avoiding both the gradient sample error and learning instability exhibited by previous methods using differentiable simulation.}
  \label{fig:teaser}
  \centering
\end{figure}

An alternative, Model-Based Reinforcement Learning (MBRL), focuses on learning environmental dynamics to enhance sample efficiency and facilitate novel methods of policy optimization. Recent MBRL research has introduced innovative dynamics models and policy learning techniques, but often without extensive independent evaluation of each component \citep{hafner2019learning, hafner2023mastering, hansen2023td}. 

When a dynamics model is available, one could employ first-order methods for policy learning, deemed theoretically more efficient \citep{mohamed2020monte,berahas2022theoretical}. This approach has been investigated in model-based control, where dynamics models guide trajectory planning \citep{kabzan2019learning,kaufmann2020deep}. However, using first-order methods to learn feedback policies within typical MBRL frameworks is less explored. This paper aims to evaluate First-Order MBRL (FO-MBRL), concentrating on policy learning and utilizing differentiable simulation for dynamics modeling.

Where model-based control literature often designs bespoke models for each problem, differentiable simulation aims to create a physics engine that is fully differentiable \citep{hu2019difftaichi,freeman2021brax,heiden2021disect,xu2021end}. Thus, applying it to a different problem is similar to using a different definition of the environment in the simulation setup (e.g., joints and links) and leaving the physics to be calculated by the engine. Short Horizon Actor-Critic (SHAC) \citep{xu2022accelerated} is an FO-MBRL approach leveraging differentiable simulation and the popular actor-critic paradigm \citep{konda1999actor}. The actor is trained in a first-order fashion, while the critic is trained model-free. This allows SHAC to learn through the highly non-convex landscape by using the critic as a smooth surrogate of the cumulative reward. While SHAC demonstrates impressive sample efficiency, it also faces challenges such as brittleness, learning instability, and dependency on extensive hyper-parameter tuning.

This study addresses these issues, shifting the focus from sample efficiency to the asymptotic performance of FO-MBRL methods in massively parallel differentiable simulation. Our analysis indicates that first-order methods suffer from significant sampling error in gradient estimation, primarily due to high dynamical gradients from stiff contact approximation ($\norm{\nabla f(\vs, \va)} \gg 0$) \citep{suh2022differentiable,lee2023differentiable}, leading to inefficiency and suboptimal policies. To address this, we propose Adaptive Horizon Actor-Critic (AHAC), a FO-MBRL algorithm that adjusts its trajectory rollout horizon to circumvent stiff dynamics (Figure \ref{fig:teaser}). Experimentally, our method shows superior asymptotic performance over MFRL baselines in complex locomotion tasks, achieving up to 64\% higher reward even when baselines are given $10^6$ times more training data. Further, AHAC's efficient use of first-order gradients enables scaling to high-dimensional motor control tasks with $152$ action dimensions.

\section{Preliminaries}
This study focuses on discrete-time and finite-horizon reinforcement learning scenarios characterized by system states $\vs \in \R^n$, actions $\va \in \R^m$, and deterministic dynamics described by the function $f: \R^n \times \R^m \rightarrow \R^n$. Actions at each timestep $t$ are sampled from a tanh-transformed stochastic policy $\va_t \sim \pi_{\vtheta} (\cdot | \vs_t)$, parameterized by $\vtheta \in \R^d$, and yield rewards from $r : \R^n \times \R^m \rightarrow \R$. The H-step return is defined as:
\begin{align*}
    R_H(\vs_1, \vtheta) = \sum_{h=1}^H r(\vs_h, \va_h) \\
    s.t. \quad \vs_{h+1} = f(\vs_h, \va_h) \quad \va_h \sim \pi_\vtheta (\cdot | \vs_h)
\end{align*}
The policy's objective is to maximize the cumulative reward:
\begin{align}
    \label{eq:objective}
    \max_\vtheta J(\vtheta) := \max_\vtheta \E[\substack{\vs_1 \sim \rho \\ \va_h \sim \pi(\cdot | \vs_h)}]{R_H(\vs_1)}
\end{align}
where $\rho$ is the initial state distribution. Without loss of generality, we simplify our derivations:

\begin{assumption}
    \label{ass:dirac-delta}
    $\rho$ is a dirac-delta distribution.
\end{assumption}

Similar to prior work \citet{duchi2012randomized,berahas2022theoretical,suh2022differentiable}, we are trying to exploit the smoothing properties of stochastic optimization on the landscape of our optimization objective. Following recent successful deep-learning approaches to MFRL \citep{schulman2017proximal, haarnoja2018soft}, we assume that our policy is stochastic, parameterized by $\vtheta$ and expressed as $\pi_\vtheta(\cdot | \vs)$.

To address the main optimization problem in Equation \ref{eq:objective}, we consider stochastic gradient estimates of  $J(\vtheta)$ using zero-order and first-order methods. To guarantee the existence of $\nabla J(\vtheta)$, we need to make certain assumptions:
\begin{definition}
    A function $g : \R^d \rightarrow \R^d$ has \textit{polynomial growth} if there exists constants $a,b$ such that $\forall \textbf{z} \in \R^d$, $||g(\textbf{z})|| \leq a(1+ ||\textbf{z}||^b)$.
\end{definition}
\begin{assumption}
    \label{ass:cont-policy}
    To ensure gradients are well defined, we assume that the policy $\pi_\vtheta ( \cdot | \vs)$ is continuously differentiable $\forall \vs \in \R^n, \forall \vtheta \in \R^d$. Furthermore, the system dynamics $f$ and reward $r$ have polynomial growth.
\end{assumption}

\subsection{Zeroth-Order Batch Gradient (ZOBG) estimates}
These weak assumptions are sufficient to make $J(\vtheta)$ differentiable in expectation by taking samples of the function value in a zeroth-order fashion \citep{williams1992simple}. This gives estimates of $\nabla J(\vtheta)$ via the stochasticity introduced by $\pi$, as first shown in \citep{williams1992simple}, and commonly referred to as as the \textit{Policy Gradient Theorem} \citep{sutton1999policy}.

\begin{definition}
    Given a sample of the H-step return $R_H(\vs_1) = \sum_{h=1}^{H} r(\vs_h, \va_h)$ following the policy $\pi$, we can estimate zero-order policy gradients via:
    \begin{align}
        \label{eq:zero-order-polcy-grad}
        \nabla_\vtheta ^{[0]} J(\vtheta) &:= \E[\va_h \sim \pi_\vtheta(\cdot | \vs_h)]{R_H (\vs_1) \sum_{h=1}^H \nabla_\vtheta \log \pi_\vtheta(\va_h | \vs_h)}
    \end{align}
\end{definition}

\begin{lemma}
    Under Assumptions \ref{ass:dirac-delta} and \ref{ass:cont-policy}, the ZOBG is an unbiased estimator of the stochastic objective $\E{\Bar{\nabla}^{[0]} J(\vtheta)} = \nabla J (\vtheta)$ where $\bar{\nabla}^{[0]} J(\vtheta)$ is the sample mean of $N$ Monte Carlo estimates of Eq. \ref{eq:zero-order-polcy-grad}.
\end{lemma}
These zero-order policy gradients are known to have high variance \citep{mohamed2020monte}, and one way to reduce their variance is by subtracting a baseline from the function estimates. Similar to \citep{suh2022differentiable}, we subtract the return given by the noise-free policy rollout where $\big( R_H (\vs_1) - R_H^*(\vs_1) \big)$ is used instead of $R_H(\vs_1)$ in Eq.~\ref{eq:zero-order-polcy-grad}.

\subsection{First-Order Batch Gradient (FOBG) estimates}
Given access to a differentiable simulator, first-order gradients induced by the policy $\pi$ can be computed via:
\begin{align}
    \label{eq:first-order-grads}
    \nabla_{\vtheta}^{[1]} J(\vtheta) := \E[\va_h \sim \pi_\vtheta(\cdot | \vs_h)]{ \nabla_\vtheta R_H(\vs_1)}
\end{align}
However, for these gradients to be well-defined, we need to make further assumptions:
\begin{assumption}
    \label{ass:smoothness}
    The dynamics $f(\vs, \va)$ and the reward $r(\vs, \va)$ are continuously differentiable $\forall \vs \in \R^n, \forall \va \in \R^m$.
\end{assumption}

Although these assumptions are necessary for the analysis of the next section, we relax them in our experiments section and consider contemporary benchmarks.

\section{Policy learning through contact}
\label{sec:toy_example}

\begin{figure}[!t]
    \centering
    \includegraphics[width=\linewidth]{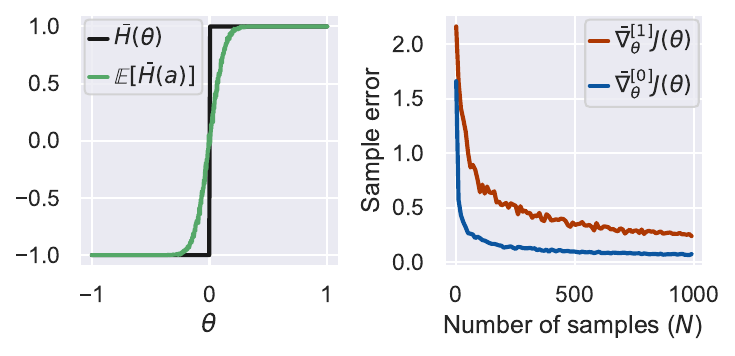}
    \caption{The left figure shows the \textbf{Soft Heaviside of Eq \ref{eq:soft-heaviside}}. The right figure shows the gradient sample error. We observe that FOBG estimates with finite $N$ exhibit a higher sample error.}
    \label{fig:soft_heaviside}
\end{figure}

Prior research has established that first-order gradients are statistically unbiased \citep{schulman2015gradient}. However, the sample error under finite $N$ is heavily dependent on the function they are trying to approximate, referred to as "empirical bias" \citep{suh2022differentiable,lee2023differentiable}. This paper explores this sampling error using the soft Heaviside function, an approximation of the Coulomb friction model, which is pivotal in discontinuous function analysis within physics simulations:
\begin{align}
\label{eq:soft-heaviside}
    \bar{H}(x) =
    \begin{cases}
        1 & x > \nu/2 \\
        2x/\nu & |x| \leq \nu/2 \\
        -1 & x < -\nu/2
    \end{cases}
\end{align}
where $a \sim \pi_\theta(\cdot) = \theta + w $ and $w \sim \mathcal{N}(0, \sigma^2)$. As shown in Appendix \ref{app:heaviside}, $\E[\pi]{\bar{H}(a)}$ is a sum of error functions whose derivative $\nabla_\theta \E[\pi]{\bar{H}(a)} \neq 0$ at $\theta=0$. However, using FOBG, we obtain $\nabla_\theta \bar{H}(a) = 0$ in samples where $|a| > \nu/2$, which occurs with probability at least $\nu / \sigma \sqrt{2\pi}$. Since in practice we are limited in sample size, this translates to sampling error that is inversely proportional to sample size, as shown in Figure \ref{fig:soft_heaviside}. Notably, when $\nu \rightarrow 0$, we achieve a more accurate approximation of the underlying discontinuous function, but we also increase the likelihood of obtaining erroneous FOBG, thus amplifying error in stochastic scenarios. We use this particular example as the differentiable simulator used in our experiments is based on the Coulomb friction model.
\\
\begin{definition}
    FOBGs exhibit sampling error relative to ZOBGs under finite samples, denoted as $B$:
    $$B = \norm{\bar{\nabla}_\vtheta^{[1]} J(\vtheta) - \bar{\nabla}_\vtheta^{[0]} J(\vtheta)}$$
\end{definition}
We analyze $B$ from the perspective of bias and variance to derive a practical upper bound:

\begin{lemma}
    \label{lem:bias-bound}
    For an H-step stochastic optimization problem under Assumptions \ref{ass:smoothness}, which also has Lipshitz-smooth policies $\| \nabla \pi_\vtheta(\va | \vs) \| \leq B_\pi$, Lipshitz-smooth reward function in both arguments $\| \nabla r(\vs, \va) \| \leq B_r$ and Lipshitz-smooth dynamics in both arguments  $\| \nabla f(\vs, \va) \| \leq B_f$ $\forall \vs \in \R^n; \va \in \R^m; \vtheta \in \R^d$, then ZOBGs remain consistently unbiased. However, FOBGs exhibit sample error bounded by:
    \begin{align}
        \label{eq:lemma}
        B \leq H B_r B_\pi \big( \dfrac{1}{2} + B_f^{H-1} \big)
    \end{align}
The proof can be found in Appendix \ref{app:lemma-proof}
\end{lemma}

\begin{figure}[!t]
    \centering
    \includegraphics[width=\linewidth]{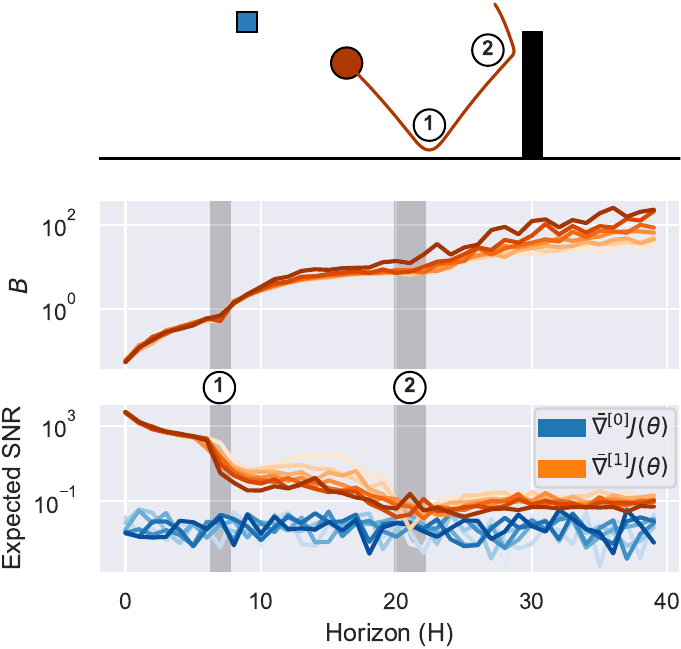}
    \caption{\textbf{Toy example where a ball is shot against a wall} trying to reach the target position in blue. The bottom two figures show gradient sample error and Expected SNR estimation with $N=1024$ samples. Darker shades designate point of contact, which negatively impact FOBG error. Higher ESNR leads to more informative gradients.}
    \label{fig:ball_grads}
\end{figure}

As $r(s,a)$ and $\pi$ are often design decisions in problems, we can create them to satisfy the assumptions laid out above. However, bounding the dynamics $||\nabla f(\vs_t, \va_t)||$ is impossible due to the natural discontinuities of physics \citep{lee2023differentiable} leading to $B_f \gg B_r$ and $B_f \gg B_\pi$. This combined with the $H$ terms lead the two conclusions from Lemma \ref{lem:bias-bound}: (1) long-horizon rollout lead to increased FOBG sample error and (2) prolonged stiff contact has compound effects on FOBG sample error.

\textbf{Empirical evaluation} of Lemma \ref{lem:bias-bound}. We designed a simple experimental setup involving a ball rebounding off a wall to reach a target, as illustrated in Figure \ref{fig:ball_grads}. The initial position $\vs_1 = [x_1, y_1]$ and velocity of the ball are fixed. The objective is for the policy to learn the optimal initial orientation $\theta$ in order to reach a target position $\vs_T$ at the end, defined as $R_H(\vs_1) = \norm{\vs_H - \vs_T}_2^{-1}$. We use the additive Gaussian policy $a = \theta + w$, where $w \sim \mathcal{N}(0, \sigma^2)$. With this, zero-order gradients from Eq. \ref{eq:zero-order-polcy-grad} can be expressed as:
\begin{align*}
    % \label{eq:zero-order-polcy-grad}
    \nabla_\vtheta ^{[0]} J(\theta) \approx \dfrac{1}{N \sigma^2} \sum_{i=1}^{N} \big( R_H^{(i)}(\vs_1) - R_H^{*(i)}(\vs_1) \big) w^{(i)}
\end{align*}
We collect $N=1024$ samples of each gradient type for each timestep with $H=40$. Figure \ref{fig:ball_grads} shows that the sample error remains low until the ball encounters contact, after which it starts growing, validating our proposed lemma. Additionally, the error also affects the gradient variance, where ZOBG follow $\Var{\nabla^{[0]}J(\theta)} \leq \sigma^{-2}H B_r^2 B_\pi^2$ \citep{suh2022differentiable}. However, FOBG variance behaves similarly to Lemma \ref{lem:bias-bound}, growing exponentially after contact. In Figure \ref{fig:ball_grads}, instead of variance, we show Expected SNR (Eq. \ref{eq:esnr}) as proposed by \citep{parmas2023model}, with higher values translating to more informative gradients. These results suggest that FOBGs exhibit sample error under contact dynamics, which is further worsened with long trajectories \citep{lee2023differentiable,zhong2023improving}.
\vspace{-4mm}
\begin{align} \label{eq:esnr}
    \operatorname{ESNR}(\nabla J(\vtheta)) = \E{\dfrac{\sum \E{\nabla J (\vtheta)}^2}{\sum \Var{\nabla J (\vtheta)}}}
\end{align}
\vspace{-6mm}

\section{Adaptive Horizon Actor-Critic (AHAC)}

\begin{figure}[!t]
    % \vspace{-0.5cm}
    \centering
    \includegraphics[width=0.98\linewidth]{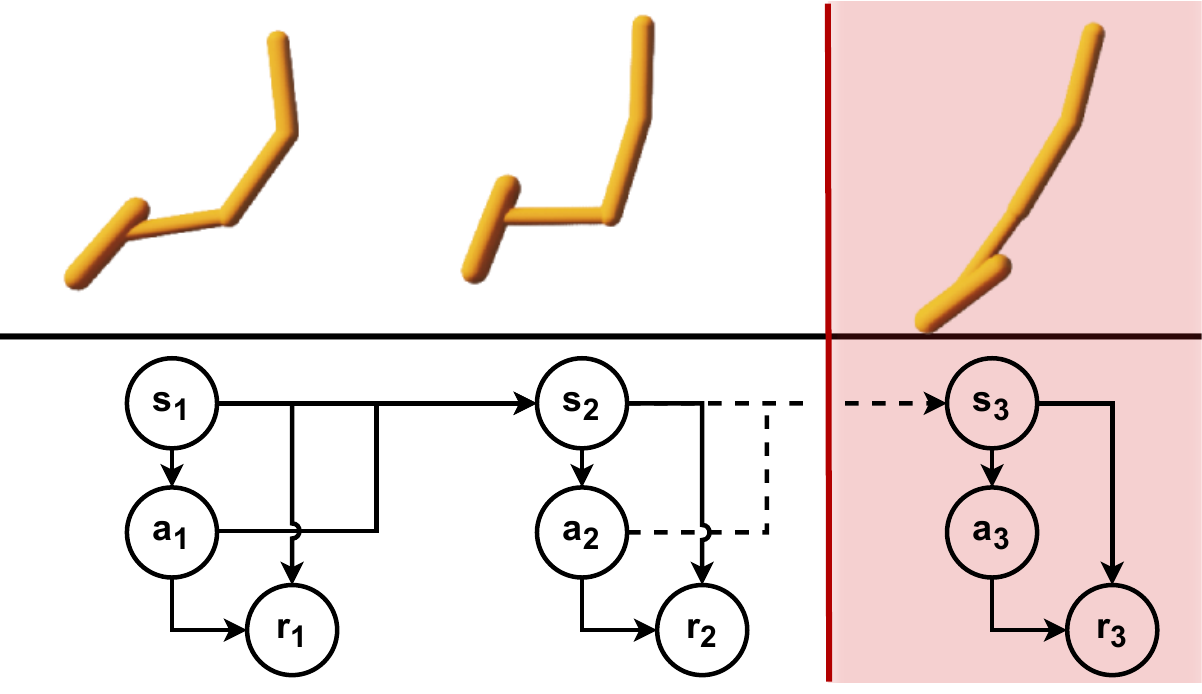}
    \caption{\textbf{Example $H=3$ step trajectory} where $\vs_3$ is in contact at which point the trajectory is truncated. When optimizing this trajectory, we completely omit the stiff dynamics gradient $\nabla f(\vs_2, \va_2)$ leading to stabler and less erroneous FOBGs.}
    \label{fig:rollout}
\end{figure}

\subsection{Learning through contact in a single environment} \label{sec:ahac-1}
With a clearer understanding of the influence of stiff contact, we aim to develop a First-Order MBRL approach for contact-rich continuous control tasks. Unlike the toy example of the previous section, standard RL multi-step decision processes allow for the avoidance of stiff dynamics gradients using  \textit{contact truncation}.
Consider the example shown in Figure \ref{fig:rollout}. 
Truncating the trajectory at the point of contact yields reward gradients without the gradient of stiff dynamics (striked out in red):
\vspace{-3mm}
\begin{align*}
    \nabla_\vtheta r(\vs_3, \va_3) = &\nabla_{\va_3} r(\vs_3, \va_3) \nabla_\vtheta \pi_\vtheta(\va_3 | \vs_3) \\ &\quad + \hcancel[red]{\nabla_{\vs_3}r(\vs_3, \va_3) \nabla_\vtheta f(\vs_2, \va_2)}
\end{align*}
% \vspace{-3mm}

\begin{figure}[!t]
    \centering
    \includegraphics[width=0.95\linewidth]{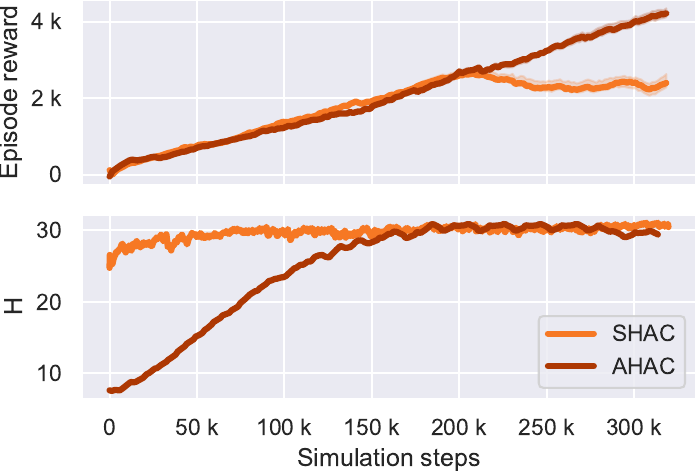}
    \vspace{-5pt}
    \caption{\textbf{Comparison between SHAC and AHAC-1 on the Hopper task with only a single environment}. The figure shows rewards and horizons achieved over 5 different random seeds, with the 50\% IQM plotted. Note that both algorithms have some horizon oscillation due to the early termination mechanism of the simulator, as noted in Appendix \ref{app:env-details}.}
    \label{fig:ahac1}
\end{figure}

We introduce an FO-MBRL algorithm with an actor-critic architecture, akin to SHAC \citep{xu2022accelerated}. The critic, denoted as $V_\psi(\vs)$, is model-free and trained using TD($\lambda$) \citep{sutton2018reinforcement} over an $H$-step horizon:

\begin{align*}
    R_h(\vs_t) := \sum_{n=t}^{t+h-1} \gamma^{n-t} r(\vs_n, \va_n) + \gamma^{t+h} V_\vpsi (\vs_{t+h}) \\
    \hat{V}(\vs_t) := (1-\lambda) \bigg[ \sum_{h=1}^{H-t-1} \lambda^{h-1} R_h(\vs_t) \bigg] + \lambda^{H-t-1} R_H(\vs_t)
\end{align*}
The critic loss becomes $\mathcal{L}_V(\vpsi)$, while the actor is trained using FOBG as in Equation \ref{eq:first-order-grads}, with the addition of the critic value estimate:
\begin{equation} \label{eq:critic-update}
    \mathcal{L}_V(\vpsi) := \sum_{h=t}^{t+H} \norm{V_\vpsi(\vs_h) - \hat{V}(\vs_h)}_2^2
\end{equation}
\vspace{-10pt}
\begin{equation} \label{eq:actor-update}
    J(\vtheta) := \sum_{h=t}^{t+H-1} \gamma^{h-t} r(\vs_h, \va_h) + \gamma^H V_\vpsi(\vs_{t+T})
\end{equation}
Unlike fixed-horizon model-based rollouts in \citep{xu2022accelerated}, our policy is rolled out until stiff contact is detected in simulation, leading to a dynamic horizon adjustment to prevent gradient explosion. However, not all contact results in high error; therefore, we truncate only on stiff contact $\norm{\nabla f(\vs_t, \va_t)} > C$, where $C$ is the contact stiffness threshold. We refer to this algorithm as Adaptive Horizon Actor-Critic 1 (AHAC-1) (see Appendix \ref{app:ahac-1}).

AHAC-1's performance was tested in a toy locomotion environment. We re-implement the popular Hopper task, where a single-legged agent hops in one axis and is rewarded for high forward velocity (Figure \ref{fig:teaser}). Compared to SHAC, which employs a fixed horizon of $H=32$, AHAC-1 adjusts its horizon based on a contact stiffness threshold of $C=500$. Results in Figure \ref{fig:ahac1} indicate that AHAC-1 achieves a higher reward than SHAC. We believe that the more erroneous SHAC gradients steer it towards local minima, while our proposed approach manages to circumvent them and achieve a higher asymptotic reward. However, AHAC-1 is not applicable to parallel vectorized environments due to the challenge of asynchronous trajectory truncation, which leads to infinitely long compute graphs.

\begin{figure}[!t]
    \centering
    \includegraphics[width=\linewidth]{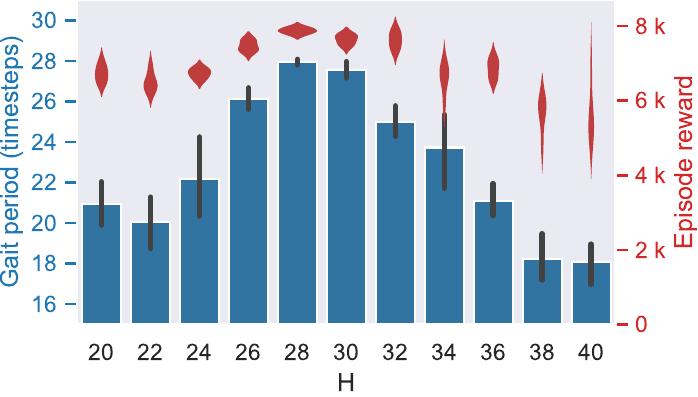}
    \vspace{-20pt}
    \caption{\textbf{An ablation of short horizons $H$} for the SHAC algorithm applied to Ant. Each run is trained until convergence for 5 seeds. The reward peaked and exhibited the least variance when the horizon length approximated the optimal gait period $H \approx 28$.}
    \label{fig:ant_gaits}
\end{figure}

\subsection{Scaling learning with synchronous parallelization}
To address the issue of asynchronous truncation, we explored the short-horizon methodology of SHAC, incorporating graph truncation at stiff contacts. However, this method did not improve performance, likely due to gradient variances across differing trajectory lengths. Consequently, our research pivoted to examine the effect of horizon length H on policy optimality, especially in contact-based tasks like locomotion that demand specific gait patterns.

We conducted empirical tests using the SHAC algorithm on the Ant locomotion task, where the goal for a quadruped robot is to maximize forward velocity. By altering the horizon length H in SHAC, our findings in Figure \ref{fig:ant_gaits} reveal a correlation between gait period and horizon length, with optimal performance at $H=28$.

\textbf{Two key insights} emerged from this study: (1) each task possesses an inherent optimal model-based horizon length $H$, closely linked to the gait period; (2) the optimal horizon correlates with the highest reward and lowest variance, aligning with the findings of Lemma \ref{lem:bias-bound}. These insights informed the development of a generalized, GPU-parallelized version of AHAC-1, termed \textit{AHAC}. While retaining the same critic training methodology as outlined in Equation \ref{eq:critic-update}, AHAC introduces a novel constrained objective for the actor.
\vspace{-0.25cm}
\begin{align} \label{eq:actor-update-constrained}
\begin{split}
    J(\vtheta) := &\sum_{h=t}^{t+H-1} \gamma^{h-t} r(\vs_h, \va_h) + \gamma^H V_\vpsi(\vs_{t+H}) \\ &s.t. \quad \| \nabla f(\vs_t, \va_t) \| \leq C \quad \forall t \in \{0, .., H\}
\end{split}
\end{align}

\begin{algorithm}[!t]
   \caption{Adaptive Horizon Actor-Critic}
   \label{alg:ahac}
\begin{algorithmic}[1]
    % \State \textbf{Given}: $\gamma$: discount rate
    \State \textbf{Given}: $\alpha_\vtheta, \alpha_\vphi, \alpha_\vpsi$: learning rates
    \State \textbf{Given}: $C$: contact threshold
    \State \textbf{Initialize learnable parameters} $\vtheta, \vpsi, H, \vphi = \textbf{0}$
    \State $t \leftarrow 0$
    % \vskip 3pt
    
    \While{episode not done}
        \State \text{Initialize rollout buffer } $D$
        \For{$h=0, 1, .., H$} \Comment{rollout policy}
            \State $\va_{t+h} \sim \pi_\vtheta(\cdot | \vs_{t+h})$
            \State $r_{t+h} = r(\vs_{t+h}, \va_{t+h})$
            \State $\vs_{t+h+1} = f(\vs_{t+h}, \va_{t+h})$
            \State $D \leftarrow D \cup \{(\vs_{t+h}, \va_{t+h}, r_{t+h}, V_\vpsi(\vs_{t+h+1}))\}$
        \EndFor
        \vskip 3pt
        
        \State $\vtheta \leftarrow \vtheta + \alpha_\vtheta \nabla_\vtheta \mathcal{L}_\pi (\vtheta, \vphi) $ \Comment{train actor (Eq. \ref{eq:lagrangian})}
        \State $\vphi \leftarrow \vphi - \alpha_\vphi \nabla_\vphi \mathcal{L}_\pi (\vtheta, \vphi)$
        \State $H \leftarrow H + \alpha_\vphi \sum_{h=0}^H \phi_h$
        \vskip 3pt
        
        \While{not converged} \Comment{train critic (Eq. \ref{eq:critic-update})}
            \State sample $(\vs, \hat{V}(\vs)) \sim D$
            \State $\vpsi \leftarrow \vpsi - \alpha_\vpsi \nabla_\vpsi \mathcal{L}_V(\vpsi)$
        \EndWhile
        \State $t \leftarrow t  + H$
    \EndWhile
\end{algorithmic}
\end{algorithm}

The objective seeks to maximize the reward while ensuring that all contact stiffness remains below the $C$ threshold. Using the Lagrangian formulation, we derive the dual problem:
\vspace{-0.5cm}
\begin{align} \label{eq:lagrangian}
\begin{split}
    \mathcal{L}_\pi (\vtheta, \vphi) &= \sum_{h=t}^{t+H-1} \gamma^{h-t} r(\vs_h, \va_h) + \gamma^H V_\vpsi(\vs_{t+H}) \\ &+ \vphi^T \left(C - \begin{bmatrix}
        \| \nabla f(\vs_t, \va_t) \| \\ \vdots \\ \| \nabla f(\vs_{t+H}, \va_{t+H}) \| 
    \end{bmatrix}\right)
\end{split}
\end{align}
By definition, $\phi_i = 0$ if the constraint is met and $\phi_i > 0$ otherwise. Thus, $\vphi$ is used to adapt the horizon, resulting in the full AHAC shown in Algorithm \ref{alg:ahac}. Additionally, we introduce a double critic that is trained until convergence, defined as a small change in the last 5 critic training iterations, $\sum_{i=n-5}^n \mathcal{L}(\vpsi) < 0.2$, where we take mini-batch samples from the rollout buffer $(\vs, \hat{V}(\vs)) \sim D$.

\begin{figure*}[ht]
     \centering
     \begin{subfigure}
         \centering
         \includegraphics[width=0.19\textwidth]{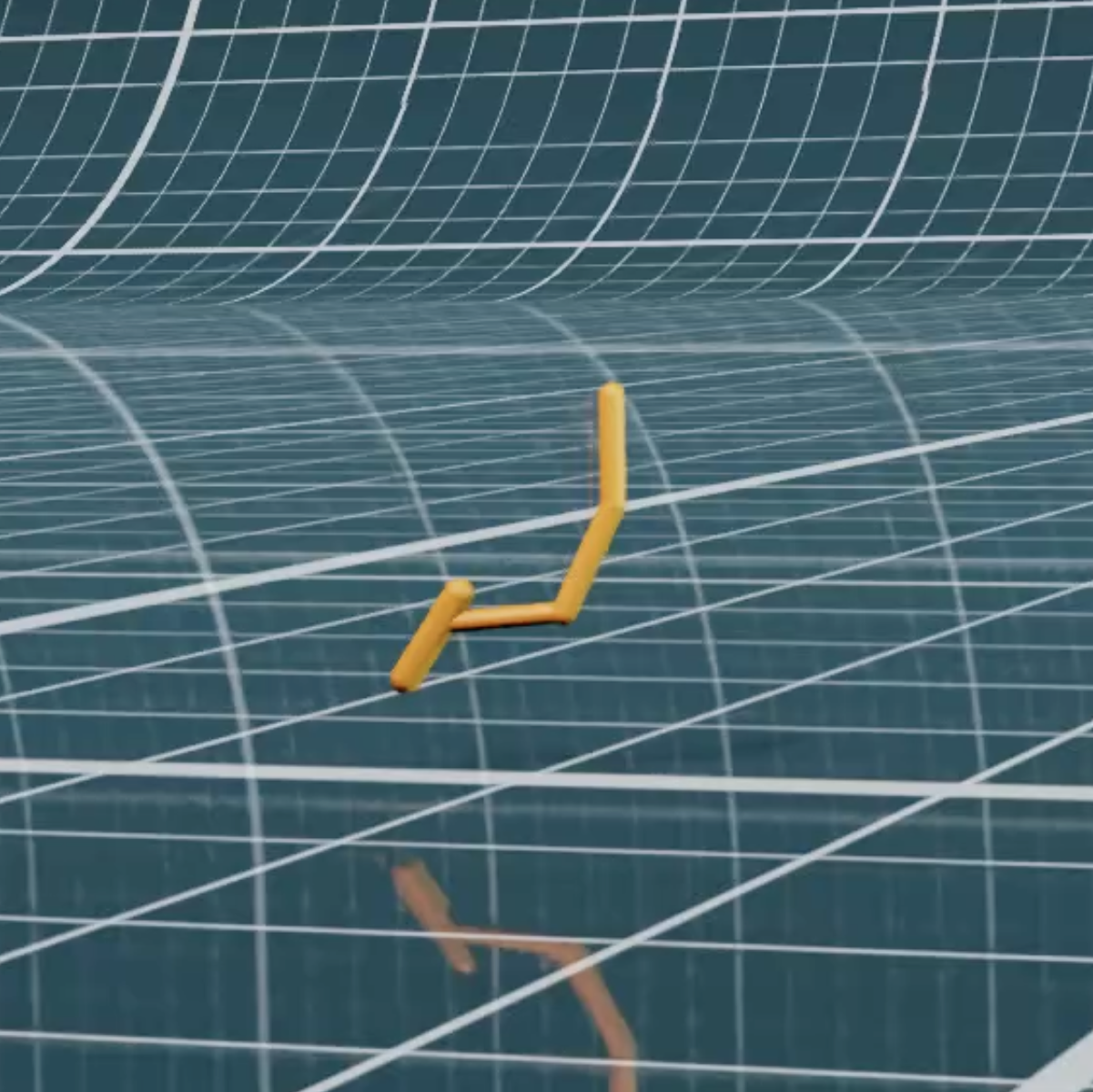}
     \end{subfigure}
     \hfill
     \begin{subfigure}
         \centering
         \includegraphics[width=0.19\textwidth]{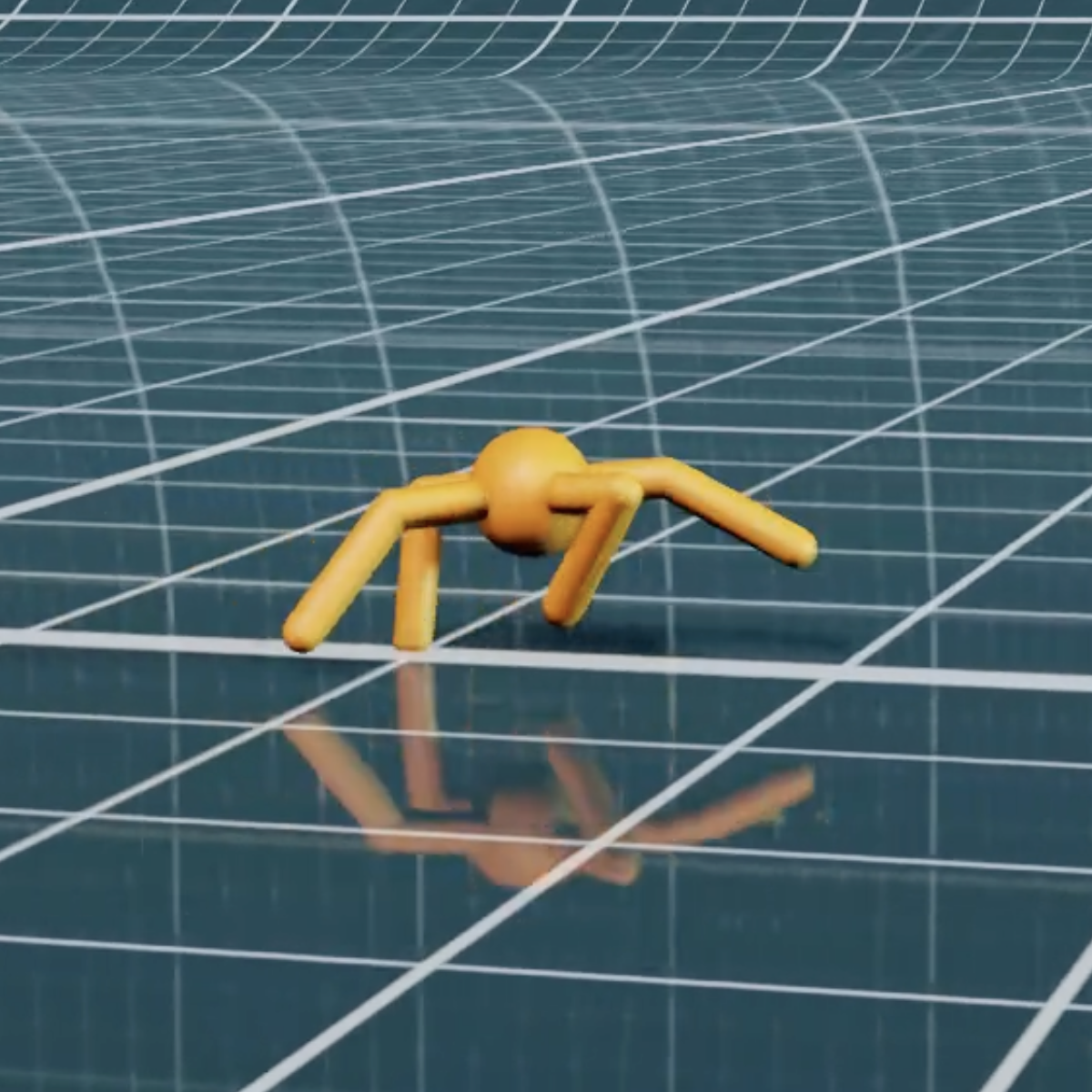}
     \end{subfigure}
     \hfill
     \begin{subfigure}
         \centering
         \includegraphics[width=0.19\textwidth]{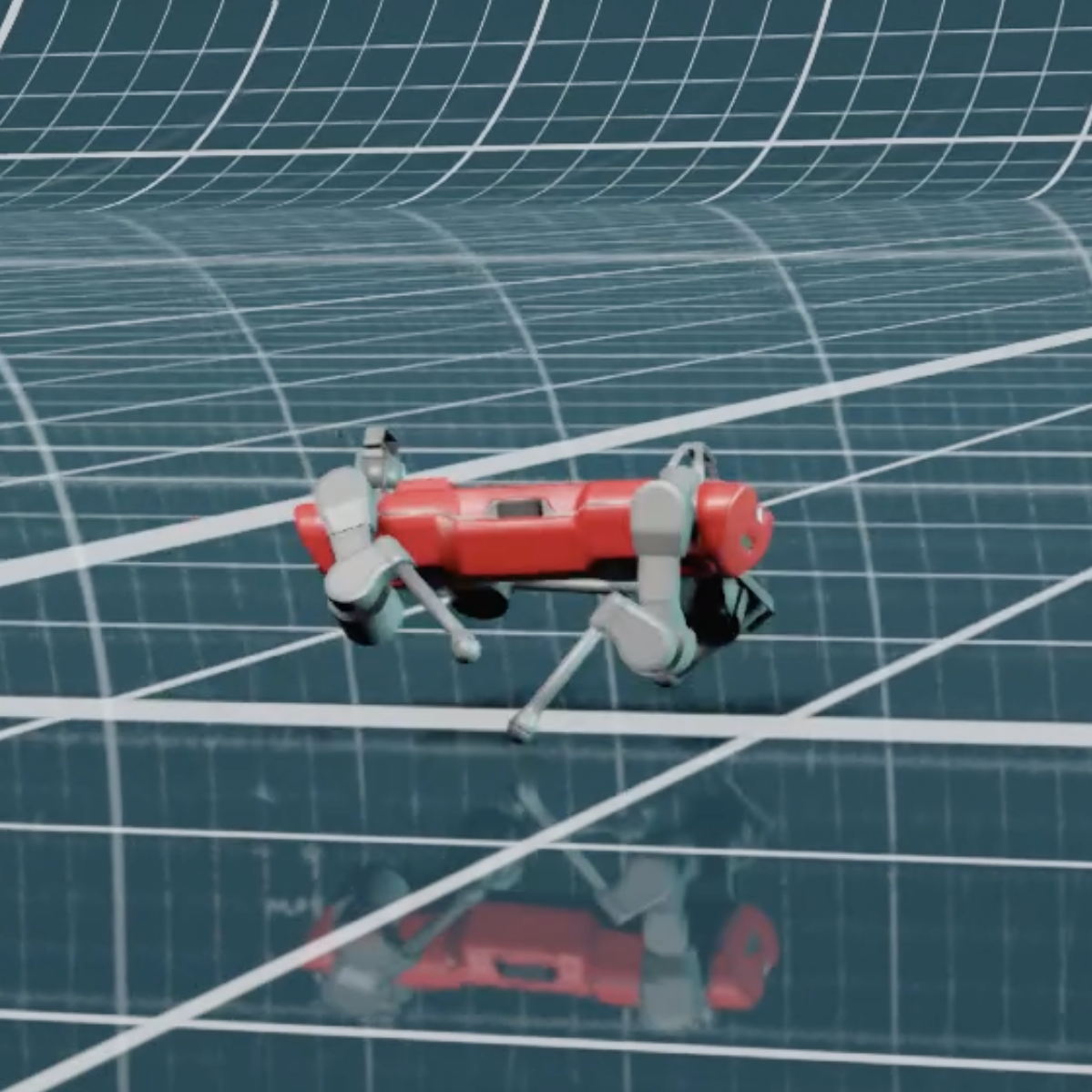}
     \end{subfigure}
     \hfill
     \begin{subfigure}
         \centering
         \includegraphics[width=.19\textwidth]{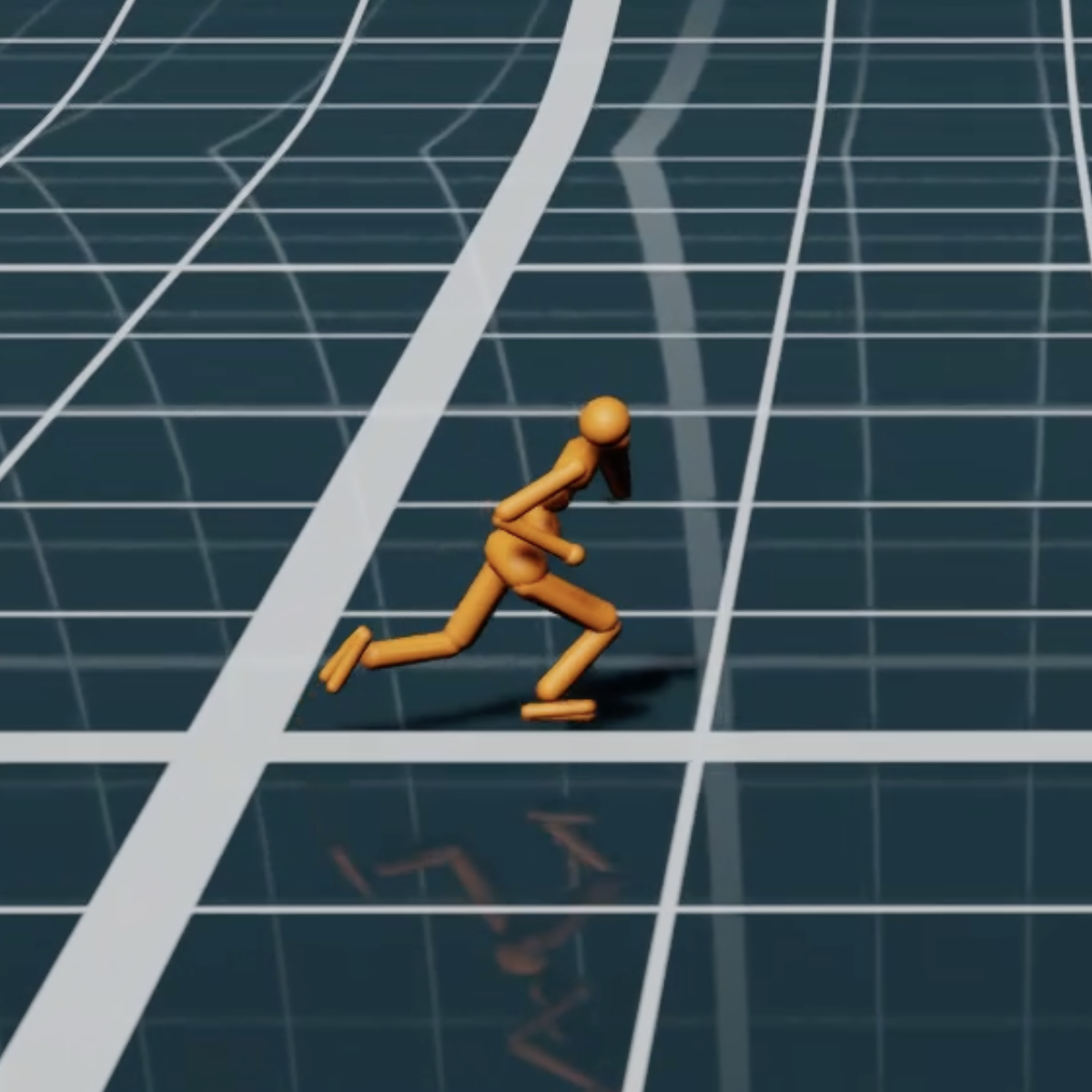}
     \end{subfigure}
         \hfill
     \begin{subfigure}
         \centering
         \includegraphics[width=0.19\textwidth]{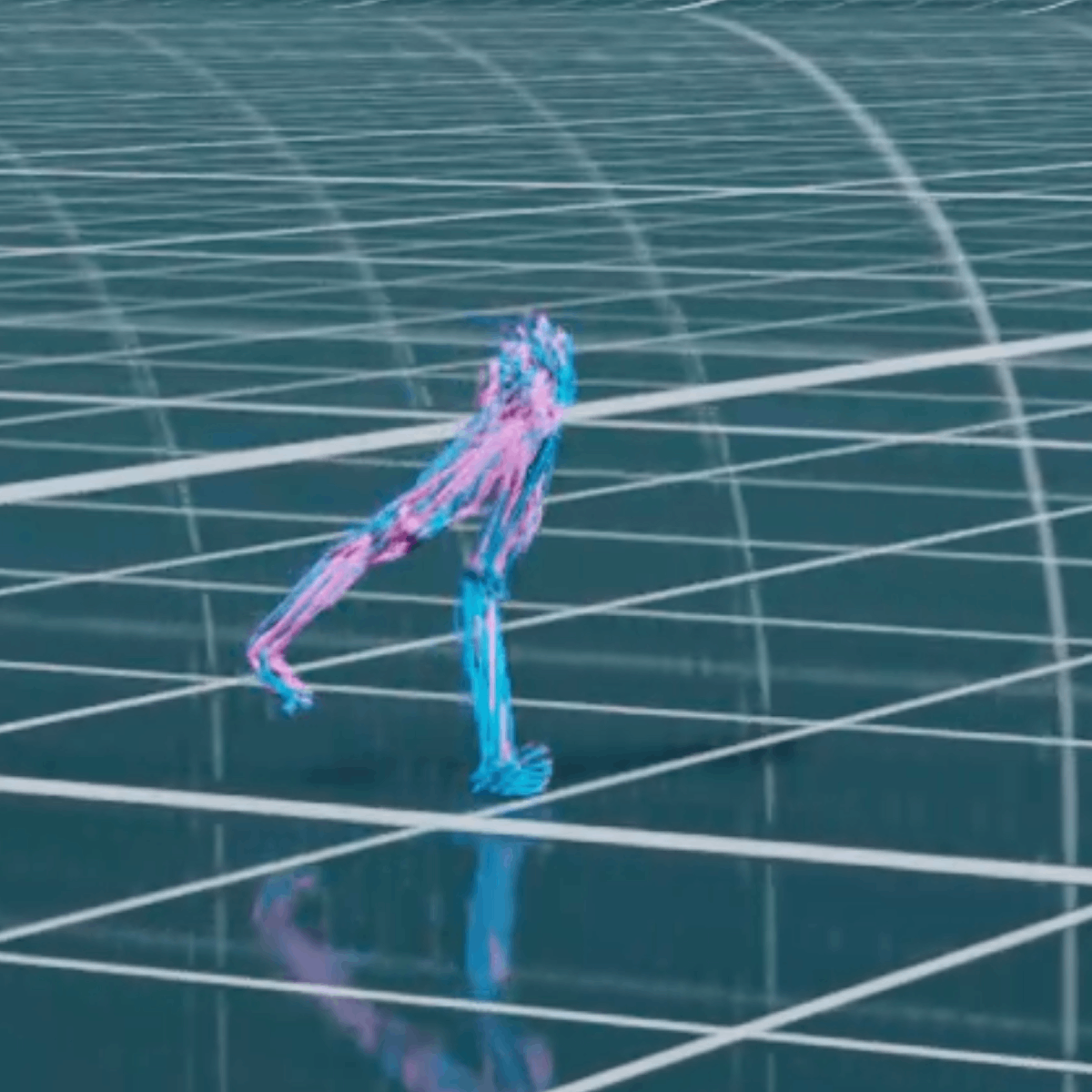}
     \end{subfigure}
        \caption{\textbf{Experimental Environments}.
        Locomotion tasks in increasing order of action space ($m$) dimensions (left to right): Hopper ($m=3$), Ant ($m=8$), Anymal ($m=12$), Humanoid ($m=21$) and SNU Humanoid ($m=152$).}
        \label{fig:main_envs_viz}
\end{figure*}

In practice, truncating based solely on the gradient of dynamics $\nabla f(\vs_t, \va_t)$ proved restrictive due to the variability of contact forces across different tasks and their evolution during the learning process. To address this, we introduced a normalization method for contact forces, utilizing modified acceleration per state dimension $\hat{\vq_t} = \max(\vq_t, 1)$ applied element-wise, resulting in normalized contact forces $\hat{\nabla} f(\vs_t, \va_t) = \diag(\hat{\vq_t}) \nabla f(\vs_t, \va_t)$. \textit{Notably, this allows using a uniform contact threshold $C$ across different tasks.}

Furthermore, considering that contact approximation forces are calculated separately in differentiable simulators, there is no need to use the full dynamics Jacobian. Instead, we employ the Jacobian, derived solely from contact forces. The differences between the SHAC and AHAC algorithms are comprehensively delineated in Appendix \ref{app:differences}.

\section{Experiments} \label{sec:experiments}

The objectives of this section are to (1) assess AHAC's ability to obtain higher asymptotic reward than MFRL baselines; (2) its efficiency in terms of wall-clock time and scalability to high-dimensional environments, and (3) identify the key contributing components of AHAC. 

\textbf{Setup.} We evaluate the proposed approach, AHAC, across a set of 5 contemporary locomotion tasks, ranging from the simpler \textit{Hopper} with $n=11$ and $m=3$, to the more complex \textit{SNU Humanoid}, which features a muscle-actuated humanoid lower body with $m=152$ (Figure \ref{fig:main_envs_viz}). All tasks aim to maximize forward velocity, chosen for its benchmark relevance \citep{tassa2018deepmind} and complex optimization landscape as alluded to by previous results \citep{haarnoja2018soft, hafner2019dream}. Experiments are based on \textit{dflex}, a differentiable rigid-body simulator with soft contact approximation, introduced by \citep{xu2022accelerated}, illustrated in Figure~\ref{fig:main_envs_viz} and described in more detail in Appendix \ref{app:simulation}. As is customary in prior work in empirical Deep RL  \citep{tassa2018deepmind}, we provide experimental results in an infinite-horizon setting and relax Assumption \ref{ass:dirac-delta}.

\textbf{Metrics.} We adopt statistical measures for a robust evaluation across 10 random seeds, utilizing the 50\% Interquartile Mean (IQM) and 95\% Confidence Interval (CI) as recommended for mitigating statistical uncertainties as suggested by \citep{agarwal2021deep}. We also report absolute as well as normalized asymptotic rewards in Appendix~\ref{app:tabular-results}.

\begin{figure}[!t]
    \centering
    \includegraphics[width=\linewidth]{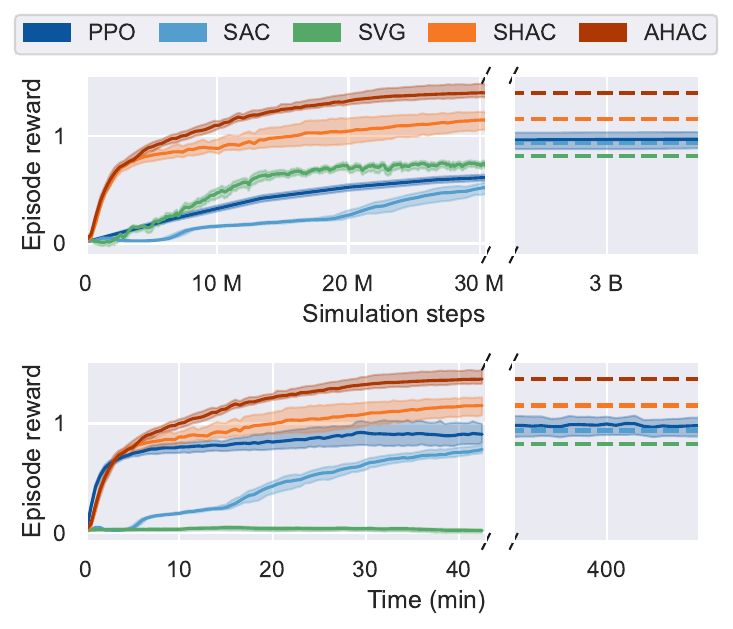}
    \vspace{-20pt}
    \caption{\textbf{Episodic rewards of the Ant task} against both simulation steps and wall clock time. The episodic reward is normalized by the highest mean reward achieved by PPO (i.e., PPO-normalized). The dashed lines represent the reward achieved by each respective algorithm at the end of their training runs.}
    \label{fig:ant_asymptotic}
    % \vspace{-0.5cm}
\end{figure}

\begin{figure*}
    \centering
    \includegraphics[width=\linewidth]{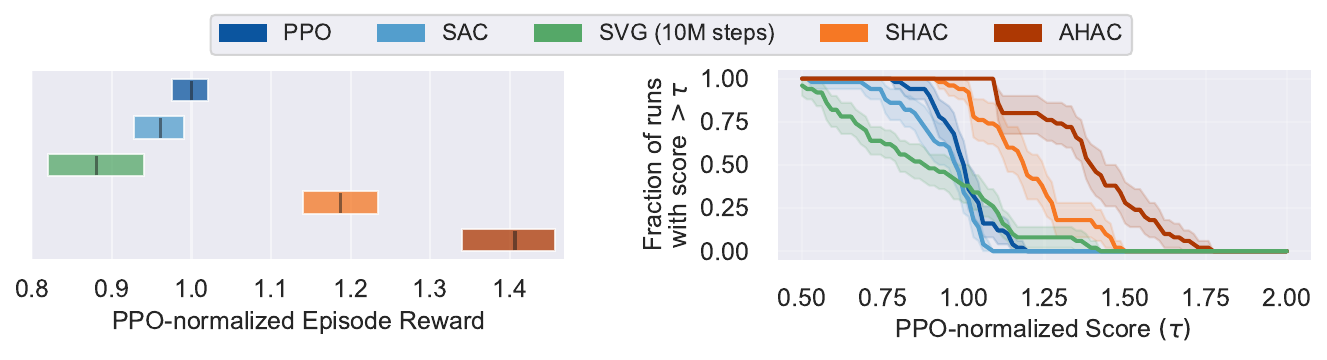}
    \vspace{-20pt}
    \caption{\textbf{Aggregate asymptotic statistics across all tasks.} The \textbf{left} figure shows 50\% IQM with 95\% CI of asymptotic episode rewards across 10 runs. We observe that AHAC is able to achieve 40\% higher reward than our best MFRL baseline, PPO. The \textbf{right} figure shows score distributions as suggested by \citep{agarwal2021deep}, which lets us understand the performance variability of each approach. Our proposed approach, AHAC, outperforms baselines even at the worst case, underlining the benefits of first-order methods.}
    \label{fig:sum_stat}
\end{figure*}

\textbf{Baselines.} This study compares first-order methods against zeroth-order methods. As such, we compare with state-of-the-art model-free methods, PPO (on-policy)  \citep{schulman2017proximal}, and SAC (off-policy) \citep{haarnoja2018soft}. For a comprehensive study, we also compare to SVG, a FO-MBRL method that does not utilize a differentiable simulator but instead learns the dynamics model\citep{amos2021model}. Additionally, we also compare our results to SHAC \citep{xu2022accelerated}, one of the best-performing methods based on differentiable simulation. We refer the reader to SHAC~\citep{xu2022accelerated} for additional comparisons with other model-based methods, which SHAC already outperforms. 
We have tuned all baselines individually to perform well per task and trained them until convergence. Due to long training times, we could not tune SVG since it was not vectorized and instead utilized the hyper-parameters from \citep{amos2021model}. All hyper-parameters are included in Appendix \ref{app:hyperparams}. For comprehension, all rewards presented in this section are normalized by the maximum reward achieved by PPO per task. We include the raw numerical results in Appendix \ref{app:tabular-results} along with further experiment details.

\begin{figure}[!t]
    \centering
    \includegraphics[width=0.95\linewidth]{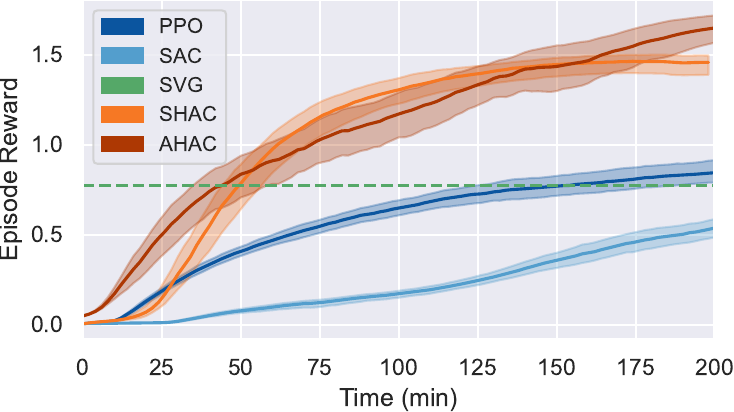}    
    \vspace{-10pt}
    \caption{\textbf{Episodic rewards of the SNU Humanoid task}, a muscle-actuated humanoid lower body with $m=152$. Results are smoothed using EWMA with $\alpha=0.9$. We observe that both SHAC and AHAC scale better to high-dimensional tasks, with the latter achieving 61\% more reward than PPO.}
    \label{fig:snu_humanoid_results}
\end{figure}

\textbf{Results.} First, we investigate the asymptotic performance of our method on the Ant task, a quadruped with symmetrical legs, $n=37$ and $m = 8$. The results in Figure~\ref{fig:ant_asymptotic} show that AHAC achieves a 41\% higher reward than the best model-free baseline, PPO, and also outperforms SHAC due to its gradient error avoidance technique. We acknowledge that MFRL methods are computationally simpler and thus also provide results against wall-clock time. Remarkably, PPO and SAC obtain worse episodic rewards over time compared to AHAC, even when they are trained for 3B timesteps, with $100\times$ more samples and $10\times$ more training time. These results suggest that even given practically infinite training data, MFRL methods cannot find truly optimal solutions due to the high variance of zeroth-order gradients.

This trend persists across all evaluated tasks, with AHAC consistently outperforming MFRL baselines. The aggregated statistics in Figure \ref{fig:sum_stat} suggest that \textit{AHAC obtains a 40\% higher reward than our main baseline, PPO}. Notably, in high-dimensional tasks such as the SNU Humanoid, AHAC's advantage becomes even more pronounced. Results from Figure \ref{fig:snu_humanoid_results} suggest that FO-MBRL methods significantly outperform MFRL baselines, with SHAC and AHAC obtaining 44\% and 64\% more reward than PPO, respectively. However, the larger confidence intervals for SHAC and AHAC hint at ongoing challenges with gradient variance associated with long rollouts. The score distributions in Figure \ref{fig:sum_stat} indicate that even with the higher variability, AHAC still outperforms MFRL baselines and exhibits better worst-case performance than SHAC. Additional results and raw metrics are provided in Appendix \ref{app:tabular-results}.

\begin{figure}[!t]
    \centering
    \includegraphics[width=0.95 \linewidth]{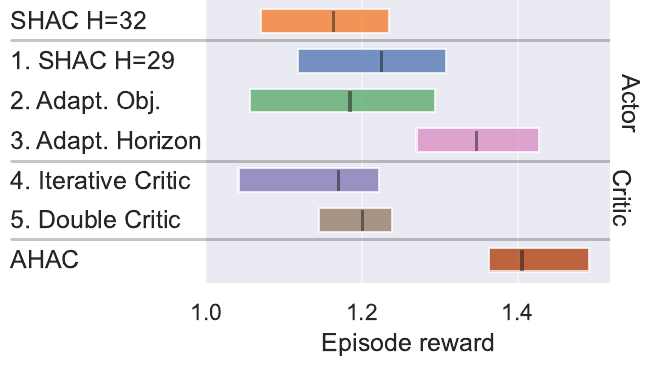}
    \vspace{-10pt}
    \caption{\textbf{Ablations of AHAC on the Ant task.} Ablating all additional introduced components reveals that the adaptive horizon objective contributes the most to improving episodic reward, while the double critic helps reduce run-to-run variance.}
    \label{fig:ablation}
\end{figure}

\begin{table*}[t]
\centering
\resizebox{0.8\textwidth}{!}{%
\begin{NiceTabular}{lccc}
\CodeBefore
  \rowcolor[HTML]{D6D6EB}{1}
  \rowcolors[HTML]{2}{eaeaf3}{FFFFFF} % seaborn colors
\Body
\toprule
    Algorithm    & Policy Learning & Value Learning & Dynamics Model \\ \midrule
    PPO \citep{schulman2017proximal}     & ZOBG & Model-Free            & -  \\
    % PlaNet \citep{hafner2019learning}  & ZOBG              & -             & Ensem. Prob. NN \\
    SAC \citep{haarnoja2018soft}     & 0-step FOBG              & Model-Free              & - \\
    MVE \citep{feinberg2018model}     & 0-step FOBG              & Model-Based           & Deterministic NN \\
    MBPO \citep{janner2019mbpo}    & 0-step FOBG              & Model-Free            & Ensemble NN \\
    PIPPS \citep{parmas2018pipps} & ZOBG \& FOBG & - & Probabilistic NN \\
                             % STEVE \citep{buckman2018sample}   & Zeroth-Order              & Model-Based             & Probabilistic NN \\
    Dreamer \citep{hafner2019dream} & FOBG              & Model-Based            & Probabilistic NN \\
    IVG \citep{byravan2020imagined}     & FOBG              & Model-Free            & Deterministic NN \\
    SVG \citep{amos2021model} & FOBG              & Model-Free            & Deterministic NN \\
    SHAC \citep{xu2022accelerated}    & FOBG              & Model-Free            & Differentiable sim. \\ 
    AHAC (ours)    & FOBG              & Model-Free            & Differentiable sim. \\ \bottomrule 
\end{NiceTabular}
}
\caption{\textbf{Comparison between RL algorithms for continuous control.} We classify methods by the policy(actor) learning approach. ZOBG stands for methods using Zeroth-Order Batch Gradients following Eq. \ref{eq:zero-order-polcy-grad}, while FOBG stands for First-Order Batch Gradient methods that differentiate through trajectories following Eq. \ref{eq:first-order-grads}. \textit{Model-Based Value Learning} refers to methods leveraging Model-Based Value Expansion (MVE) \citep{feinberg2018model}, whereas \textit{Model-Free critic learning} refers to methods using variants of TD($\lambda$) \citep{sutton2018reinforcement}.}
\label{tab:alg_compar}
\end{table*}

\textbf{Ablation study.} To elucidate the performance improvements attributed to AHAC, we dissect its pivotal modifications, outlined in Appendix \ref{app:differences}. Beginning with the SHAC baseline set at $H=32$, each ablation incrementally introduces a single modification:
\begin{enumerate}[nosep]
    \item SHAC H=29: using the $H$ converged by AHAC.
    \item Adapt. Obj.: SHAC with Eq. \ref{eq:lagrangian} and fixed $H=32$.
    \item Adapt. Horizon: SHAC with Eq. \ref{eq:lagrangian} and adapting $H$.
    \item Iterative critic: SHAC with iterative critic training.
    \item Double critic: SHAC with a double critic. 
\end{enumerate}

All experiments used the SHAC hyper-parameters, with the exception of the horizon learning rate $\alpha_\vphi$, specifically adjusted for AHAC. Notably, SHAC with an adaptive horizon (3) is equivalent to AHAC without iterative critic training and single critic implementation. Results, depicted in Figure \ref{fig:ablation}, reveal that incorporating an adaptive horizon significantly enhances the asymptotic reward. Intriguingly, adjusting to $H=29$ improves rewards over the baseline, yet does not reach the efficacy of the full adaptive horizon approach. \textit{This suggests that a static optimal horizon, even if advantageous at policy convergence, may not be optimal during training, leading to local minima.} Moreover, the double critic model notably reduces run-to-run variance, surpassing the performance stability of SHAC's single target-critic approach. Additional insights and detailed ablation results are available in Appendix \ref{app:ablation-study}.

% \newpage
\section{Related work} \label{sec:related-work}

This section reviews recent advancements in continuous control RL, adhering to the actor-critic framework \citep{konda1999actor}, where the critic appraises state-action pairs and the actor identifies optimal actions. We categorize the methods based on their policy training, value estimation, and use of a dynamics model.

In the absence of a known dynamics model, Model-Free Reinforcement Learning (MFRL) methods prevail, which enable learning of action distributions based on state information. Proximal Policy Optimization (PPO) \citep{schulman2017proximal} is an on-policy method that utilizes ZOBGs (Eq. \ref{eq:zero-order-polcy-grad}) and performs gradient updates using recent on-policy samples. Soft Actor-Critic (SAC) \citep{haarnoja2018soft} exemplifies off-policy methods which use a replay buffer to learn from any data and update the actor using 0-step FOBGs defined as $\nabla_{\vtheta} J(\vtheta) := \E[\va_h \sim \pi_\vtheta(\cdot | \vs_h)]{\nabla_\vtheta Q(\vs,\va)}$

Alternatively, Model-Based Reinforcement Learning (MBRL) methods incorporate a dynamics model to inform learning, either derived from data or assumed a priori. This model can be used to aid the critic's return estimates, which can still be trained model-free \citep{janner2019mbpo} or through back-propagatable simulated returns via Model-Based Value Expansion (MVE) \citep{feinberg2018model}. Actor training varies; it can be done using 0-step FOBGs augmented by model-generated data \citep{janner2019mbpo,feinberg2018model}. Alternatively, other work back-propagates through the dynamics model \citep{hafner2019dream,byravan2020imagined} using FOBGs (Eq. \ref{eq:first-order-grads}). \citep{parmas2018pipps} also combine ZOBGs and FOBGs, attempting to harness the best of both. Key recent work is summarized in Table \ref{tab:alg_compar}.

With the emergence of differentiable simulation, many studies \citep{hu2019chainqueen,liang2019differentiable,huang2021plasticinelab,du2021diffpd} have explored FOBG optimization by back-propagating through a dynamics model which we refer to as Back-Propagation-Through-Time (BPTT). However, BPTT faces challenges in long episodic RL tasks due to unstable gradients. \citep{xu2022accelerated} introduces Short Horizon Actor-Critic (SHAC), improving stability through a model-free critic and short rollouts, achieving performance comparable to MFRL with enhanced sample efficiency.

\section{Conclusion}

Model-free RL (MFRL) approaches are valued for their simplicity, minimal assumptions, and impressive performance. Yet, our study reveals their limitations in complex continuous control tasks, where they achieve good but sub-optimal solutions due to high gradient variance. Conversely, First-Order Model-Based RL (FO-MBRL) methods, which leverage efficient gradient propagation through dynamics, have historically lagged behind MFRL in performance.  

In this work, we analyze this issue in differentiable simulation through the scope of bias and variance. We derive Lemma \ref{lem:bias-bound} bounding the observed sample error of first-order gradients relative to zeroth-order gradients, coming to the conclusion that the source of the issue is stiff contact and long horizon rollouts. Based on these insights, we propose the Adaptive Horizon Actor-Critic (AHAC), a new FO-MBRL approach that adapts its rollout horizon during training. Our experiments show that AHAC outperforms MFRL baselines by 40\% in complex locomotion tasks, even when the latter are provided with $10^6$ times more data. Furthermore, out method maintains competitive time efficiency and shows better scalability to higher-dimensional tasks.

While AHAC outperforms MFRL methods and makes the case for first-order policy learning, it also necessitates the development of differentiable simulators. As such, we admire the simple yet capable MFRL approaches. Our work suggests that future research should not only focus on refining algorithmic approaches for policy learning but also on enhancing simulator technologies to more effectively manage gradient error. Moreover, the practical application of policies trained in differentiable simulators to real-world robotics remains a challenge.

% \clearpage
\subsection*{Impact Statement}
This paper presents work whose goal is to advance the field of Reinforcement Learning. There are many potential societal consequences of our work, none which we feel must be specifically highlighted here.

\subsection*{Acknowledgment}
The authors (IG, AG) were support by Stephen Fleming Early Career Chair as well as gifts from Nuro and Ford Motor Company, NSERC Discovery Award.
This research was supported in part through services from the Partnership for an Advanced Computing Environment (PACE) at the Georgia Institute of Technology, Atlanta, Georgia, USA.

\bibliography{main}

\begin{thebibliography}{44}
\providecommand{\natexlab}[1]{#1}
\providecommand{\url}[1]{\texttt{#1}}
\expandafter\ifx\csname urlstyle\endcsname\relax
  \providecommand{\doi}[1]{doi: #1}\else
  \providecommand{\doi}{doi: \begingroup \urlstyle{rm}\Url}\fi

\bibitem[Agarwal et~al.(2021)Agarwal, Schwarzer, Castro, Courville, and Bellemare]{agarwal2021deep}
Agarwal, R., Schwarzer, M., Castro, P.~S., Courville, A.~C., and Bellemare, M.
\newblock Deep reinforcement learning at the edge of the statistical precipice.
\newblock \emph{Advances in neural information processing systems}, 34:\penalty0 29304--29320, 2021.

\bibitem[Akkaya et~al.(2019)Akkaya, Andrychowicz, Chociej, Litwin, McGrew, Petron, Paino, Plappert, Powell, Ribas, et~al.]{akkaya2019solving}
Akkaya, I., Andrychowicz, M., Chociej, M., Litwin, M., McGrew, B., Petron, A., Paino, A., Plappert, M., Powell, G., Ribas, R., et~al.
\newblock Solving rubik's cube with a robot hand.
\newblock \emph{arXiv preprint arXiv:1910.07113}, 2019.

\bibitem[Amos et~al.(2021)Amos, Stanton, Yarats, and Wilson]{amos2021model}
Amos, B., Stanton, S., Yarats, D., and Wilson, A.~G.
\newblock On the model-based stochastic value gradient for continuous reinforcement learning.
\newblock In \emph{Learning for Dynamics and Control}, pp.\  6--20. PMLR, 2021.

\bibitem[Berahas et~al.(2022)Berahas, Cao, Choromanski, and Scheinberg]{berahas2022theoretical}
Berahas, A.~S., Cao, L., Choromanski, K., and Scheinberg, K.
\newblock A theoretical and empirical comparison of gradient approximations in derivative-free optimization.
\newblock \emph{Foundations of Computational Mathematics}, 22\penalty0 (2):\penalty0 507--560, 2022.

\bibitem[Byravan et~al.(2020)Byravan, Springenberg, Abdolmaleki, Hafner, Neunert, Lampe, Siegel, Heess, and Riedmiller]{byravan2020imagined}
Byravan, A., Springenberg, J.~T., Abdolmaleki, A., Hafner, R., Neunert, M., Lampe, T., Siegel, N., Heess, N., and Riedmiller, M.
\newblock Imagined value gradients: Model-based policy optimization with tranferable latent dynamics models.
\newblock In \emph{Conference on Robot Learning}, pp.\  566--589. PMLR, 2020.

\bibitem[Deng et~al.(2024)Deng, Park, and Ahn]{deng2024facing}
Deng, F., Park, J., and Ahn, S.
\newblock Facing off world model backbones: Rnns, transformers, and s4.
\newblock \emph{Advances in Neural Information Processing Systems}, 36, 2024.

\bibitem[Du et~al.(2021)Du, Wu, Ma, Wah, Spielberg, Rus, and Matusik]{du2021diffpd}
Du, T., Wu, K., Ma, P., Wah, S., Spielberg, A., Rus, D., and Matusik, W.
\newblock Diffpd: Differentiable projective dynamics.
\newblock \emph{ACM Transactions on Graphics (TOG)}, 41\penalty0 (2):\penalty0 1--21, 2021.

\bibitem[Duchi et~al.(2012)Duchi, Bartlett, and Wainwright]{duchi2012randomized}
Duchi, J.~C., Bartlett, P.~L., and Wainwright, M.~J.
\newblock Randomized smoothing for stochastic optimization.
\newblock \emph{SIAM Journal on Optimization}, 22\penalty0 (2):\penalty0 674--701, 2012.

\bibitem[Feinberg et~al.(2018)Feinberg, Wan, Stoica, Jordan, Gonzalez, and Levine]{feinberg2018model}
Feinberg, V., Wan, A., Stoica, I., Jordan, M.~I., Gonzalez, J.~E., and Levine, S.
\newblock Model-based value expansion for efficient model-free reinforcement learning.
\newblock In \emph{Proceedings of the 35th International Conference on Machine Learning (ICML 2018)}, 2018.

\bibitem[Freeman et~al.(2021)Freeman, Frey, Raichuk, Girgin, Mordatch, and Bachem]{freeman2021brax}
Freeman, C.~D., Frey, E., Raichuk, A., Girgin, S., Mordatch, I., and Bachem, O.
\newblock Brax--a differentiable physics engine for large scale rigid body simulation.
\newblock \emph{arXiv preprint arXiv:2106.13281}, 2021.

\bibitem[Haarnoja et~al.(2018)Haarnoja, Zhou, Hartikainen, Tucker, Ha, Tan, Kumar, Zhu, Gupta, Abbeel, et~al.]{haarnoja2018soft}
Haarnoja, T., Zhou, A., Hartikainen, K., Tucker, G., Ha, S., Tan, J., Kumar, V., Zhu, H., Gupta, A., Abbeel, P., et~al.
\newblock Soft actor-critic algorithms and applications.
\newblock \emph{arXiv preprint arXiv:1812.05905}, 2018.

\bibitem[Hafner et~al.(2019{\natexlab{a}})Hafner, Lillicrap, Ba, and Norouzi]{hafner2019dream}
Hafner, D., Lillicrap, T., Ba, J., and Norouzi, M.
\newblock Dream to control: Learning behaviors by latent imagination.
\newblock \emph{arXiv preprint arXiv:1912.01603}, 2019{\natexlab{a}}.

\bibitem[Hafner et~al.(2019{\natexlab{b}})Hafner, Lillicrap, Fischer, Villegas, Ha, Lee, and Davidson]{hafner2019learning}
Hafner, D., Lillicrap, T., Fischer, I., Villegas, R., Ha, D., Lee, H., and Davidson, J.
\newblock Learning latent dynamics for planning from pixels.
\newblock In \emph{International conference on machine learning}, pp.\  2555--2565. PMLR, 2019{\natexlab{b}}.

\bibitem[Hafner et~al.(2023)Hafner, Pasukonis, Ba, and Lillicrap]{hafner2023mastering}
Hafner, D., Pasukonis, J., Ba, J., and Lillicrap, T.
\newblock Mastering diverse domains through world models.
\newblock \emph{arXiv preprint arXiv:2301.04104}, 2023.

\bibitem[Hansen et~al.(2023)Hansen, Su, and Wang]{hansen2023td}
Hansen, N., Su, H., and Wang, X.
\newblock Td-mpc2: Scalable, robust world models for continuous control.
\newblock \emph{arXiv preprint arXiv:2310.16828}, 2023.

\bibitem[Heiden et~al.(2021)Heiden, Macklin, Narang, Fox, Garg, and Ramos]{heiden2021disect}
Heiden, E., Macklin, M., Narang, Y.~S., Fox, D., Garg, A., and Ramos, F.
\newblock {DiSECt: A Differentiable Simulation Engine for Autonomous Robotic Cutting}.
\newblock \emph{Robotics: Science and Systems}, 2021.

\bibitem[Hu et~al.(2019{\natexlab{a}})Hu, Anderson, Li, Sun, Carr, Ragan-Kelley, and Durand]{hu2019difftaichi}
Hu, Y., Anderson, L., Li, T.-M., Sun, Q., Carr, N., Ragan-Kelley, J., and Durand, F.
\newblock Difftaichi: Differentiable programming for physical simulation.
\newblock \emph{arXiv preprint arXiv:1910.00935}, 2019{\natexlab{a}}.

\bibitem[Hu et~al.(2019{\natexlab{b}})Hu, Liu, Spielberg, Tenenbaum, Freeman, Wu, Rus, and Matusik]{hu2019chainqueen}
Hu, Y., Liu, J., Spielberg, A., Tenenbaum, J.~B., Freeman, W.~T., Wu, J., Rus, D., and Matusik, W.
\newblock Chainqueen: A real-time differentiable physical simulator for soft robotics.
\newblock In \emph{2019 International conference on robotics and automation (ICRA)}, pp.\  6265--6271. IEEE, 2019{\natexlab{b}}.

\bibitem[Huang et~al.(2021)Huang, Hu, Du, Zhou, Su, Tenenbaum, and Gan]{huang2021plasticinelab}
Huang, Z., Hu, Y., Du, T., Zhou, S., Su, H., Tenenbaum, J.~B., and Gan, C.
\newblock Plasticinelab: A soft-body manipulation benchmark with differentiable physics.
\newblock \emph{arXiv preprint arXiv:2104.03311}, 2021.

\bibitem[Hutter et~al.(2016)Hutter, Gehring, Jud, Lauber, Bellicoso, Tsounis, Hwangbo, Bodie, Fankhauser, Bloesch, et~al.]{hutter2016anymal}
Hutter, M., Gehring, C., Jud, D., Lauber, A., Bellicoso, C.~D., Tsounis, V., Hwangbo, J., Bodie, K., Fankhauser, P., Bloesch, M., et~al.
\newblock Anymal-a highly mobile and dynamic quadrupedal robot.
\newblock In \emph{2016 IEEE/RSJ international conference on intelligent robots and systems (IROS)}, pp.\  38--44. IEEE, 2016.

\bibitem[Hwangbo et~al.(2017)Hwangbo, Sa, Siegwart, and Hutter]{hwangbo2017control}
Hwangbo, J., Sa, I., Siegwart, R., and Hutter, M.
\newblock Control of a quadrotor with reinforcement learning.
\newblock \emph{IEEE Robotics and Automation Letters}, 2\penalty0 (4):\penalty0 2096--2103, 2017.

\bibitem[Hwangbo et~al.(2019)Hwangbo, Lee, Dosovitskiy, Bellicoso, Tsounis, Koltun, and Hutter]{hwangbo2019learning}
Hwangbo, J., Lee, J., Dosovitskiy, A., Bellicoso, D., Tsounis, V., Koltun, V., and Hutter, M.
\newblock Learning agile and dynamic motor skills for legged robots.
\newblock \emph{Science Robotics}, 4\penalty0 (26):\penalty0 eaau5872, 2019.

\bibitem[Janner et~al.(2019)Janner, Fu, Zhang, and Levine]{janner2019mbpo}
Janner, M., Fu, J., Zhang, M., and Levine, S.
\newblock When to trust your model: Model-based policy optimization.
\newblock In \emph{Advances in Neural Information Processing Systems}, 2019.

\bibitem[Kabzan et~al.(2019)Kabzan, Hewing, Liniger, and Zeilinger]{kabzan2019learning}
Kabzan, J., Hewing, L., Liniger, A., and Zeilinger, M.~N.
\newblock Learning-based model predictive control for autonomous racing.
\newblock \emph{IEEE Robotics and Automation Letters}, 4\penalty0 (4):\penalty0 3363--3370, 2019.

\bibitem[Kaufmann et~al.(2020)Kaufmann, Loquercio, Ranftl, M{\"u}ller, Koltun, and Scaramuzza]{kaufmann2020deep}
Kaufmann, E., Loquercio, A., Ranftl, R., M{\"u}ller, M., Koltun, V., and Scaramuzza, D.
\newblock Deep drone acrobatics.
\newblock \emph{arXiv preprint arXiv:2006.05768}, 2020.

\bibitem[Konda \& Tsitsiklis(1999)Konda and Tsitsiklis]{konda1999actor}
Konda, V. and Tsitsiklis, J.
\newblock Actor-critic algorithms.
\newblock \emph{Advances in neural information processing systems}, 12, 1999.

\bibitem[Lee et~al.(2023)Lee, Lee, and Lee]{lee2023differentiable}
Lee, M., Lee, J., and Lee, D.
\newblock Differentiable dynamics simulation using invariant contact mapping and damped contact force.
\newblock In \emph{2023 IEEE International Conference on Robotics and Automation (ICRA)}, pp.\  11683--11689. IEEE, 2023.

\bibitem[Liang et~al.(2019)Liang, Lin, and Koltun]{liang2019differentiable}
Liang, J., Lin, M., and Koltun, V.
\newblock Differentiable cloth simulation for inverse problems.
\newblock \emph{Advances in Neural Information Processing Systems}, 32, 2019.

\bibitem[Lillicrap et~al.(2015)Lillicrap, Hunt, Pritzel, Heess, Erez, Tassa, Silver, and Wierstra]{lillicrap2015continuous}
Lillicrap, T.~P., Hunt, J.~J., Pritzel, A., Heess, N., Erez, T., Tassa, Y., Silver, D., and Wierstra, D.
\newblock Continuous control with deep reinforcement learning.
\newblock \emph{arXiv preprint arXiv:1509.02971}, 2015.

\bibitem[Macklin(2022)]{warp2022}
Macklin, M.
\newblock Warp: A high-performance python framework for gpu simulation and graphics.
\newblock \url{https://github.com/nvidia/warp}, March 2022.
\newblock NVIDIA GPU Technology Conference (GTC).

\bibitem[Mohamed et~al.(2020)Mohamed, Rosca, Figurnov, and Mnih]{mohamed2020monte}
Mohamed, S., Rosca, M., Figurnov, M., and Mnih, A.
\newblock Monte carlo gradient estimation in machine learning.
\newblock \emph{The Journal of Machine Learning Research}, 21\penalty0 (1):\penalty0 5183--5244, 2020.

\bibitem[Parmas et~al.(2018)Parmas, Rasmussen, Peters, and Doya]{parmas2018pipps}
Parmas, P., Rasmussen, C.~E., Peters, J., and Doya, K.
\newblock Pipps: Flexible model-based policy search robust to the curse of chaos.
\newblock In \emph{International Conference on Machine Learning}, pp.\  4065--4074. PMLR, 2018.

\bibitem[Parmas et~al.(2023)Parmas, Seno, and Aoki]{parmas2023model}
Parmas, P., Seno, T., and Aoki, Y.
\newblock Model-based reinforcement learning with scalable composite policy gradient estimators.
\newblock In \emph{International Conference on Machine Learning}, pp.\  27346--27377. PMLR, 2023.

\bibitem[Rudin et~al.(2022)Rudin, Hoeller, Reist, and Hutter]{rudin2022learning}
Rudin, N., Hoeller, D., Reist, P., and Hutter, M.
\newblock Learning to walk in minutes using massively parallel deep reinforcement learning.
\newblock In \emph{Conference on Robot Learning}, pp.\  91--100. PMLR, 2022.

\bibitem[Schulman et~al.(2015)Schulman, Heess, Weber, and Abbeel]{schulman2015gradient}
Schulman, J., Heess, N., Weber, T., and Abbeel, P.
\newblock Gradient estimation using stochastic computation graphs.
\newblock \emph{Advances in neural information processing systems}, 28, 2015.

\bibitem[Schulman et~al.(2017)Schulman, Wolski, Dhariwal, Radford, and Klimov]{schulman2017proximal}
Schulman, J., Wolski, F., Dhariwal, P., Radford, A., and Klimov, O.
\newblock Proximal policy optimization algorithms.
\newblock \emph{arXiv preprint arXiv:1707.06347}, 2017.

\bibitem[Suh et~al.(2022)Suh, Simchowitz, Zhang, and Tedrake]{suh2022differentiable}
Suh, H.~J., Simchowitz, M., Zhang, K., and Tedrake, R.
\newblock Do differentiable simulators give better policy gradients?
\newblock In \emph{International Conference on Machine Learning}, pp.\  20668--20696. PMLR, 2022.

\bibitem[Sutton \& Barto(2018)Sutton and Barto]{sutton2018reinforcement}
Sutton, R.~S. and Barto, A.~G.
\newblock \emph{Reinforcement learning: An introduction}.
\newblock MIT press, 2018.

\bibitem[Sutton et~al.(1999)Sutton, McAllester, Singh, and Mansour]{sutton1999policy}
Sutton, R.~S., McAllester, D., Singh, S., and Mansour, Y.
\newblock Policy gradient methods for reinforcement learning with function approximation.
\newblock \emph{Advances in neural information processing systems}, 12, 1999.

\bibitem[Tassa et~al.(2018)Tassa, Doron, Muldal, Erez, Li, Casas, Budden, Abdolmaleki, Merel, Lefrancq, et~al.]{tassa2018deepmind}
Tassa, Y., Doron, Y., Muldal, A., Erez, T., Li, Y., Casas, D. d.~L., Budden, D., Abdolmaleki, A., Merel, J., Lefrancq, A., et~al.
\newblock Deepmind control suite.
\newblock \emph{arXiv preprint arXiv:1801.00690}, 2018.

\bibitem[Williams(1992)]{williams1992simple}
Williams, R.~J.
\newblock Simple statistical gradient-following algorithms for connectionist reinforcement learning.
\newblock \emph{Reinforcement learning}, pp.\  5--32, 1992.

\bibitem[Xu et~al.(2021)Xu, Chen, Zlokapa, Foshey, Matusik, Sueda, and Agrawal]{xu2021end}
Xu, J., Chen, T., Zlokapa, L., Foshey, M., Matusik, W., Sueda, S., and Agrawal, P.
\newblock An end-to-end differentiable framework for contact-aware robot design.
\newblock \emph{arXiv preprint arXiv:2107.07501}, 2021.

\bibitem[Xu et~al.(2022)Xu, Makoviychuk, Narang, Ramos, Matusik, Garg, and Macklin]{xu2022accelerated}
Xu, J., Makoviychuk, V., Narang, Y., Ramos, F., Matusik, W., Garg, A., and Macklin, M.
\newblock Accelerated policy learning with parallel differentiable simulation.
\newblock \emph{arXiv preprint arXiv:2204.07137}, 2022.

\bibitem[Zhong et~al.(2023)Zhong, Han, Dey, and Brikis]{zhong2023improving}
Zhong, Y.~D., Han, J., Dey, B., and Brikis, G.~O.
\newblock Improving gradient computation for differentiable physics simulation with contacts.
\newblock In \emph{Learning for Dynamics and Control Conference}, pp.\  128--141. PMLR, 2023.

\end{thebibliography}
\bibliographystyle{icml2024}

%%%%%%%%%%%%%%%%%%%%%%%%%%%%%%%%%%%%%%%%%%%%%%%%%%%%%%%%%%%%%%%%%%%%%%%%%%%%%%%
%%%%%%%%%%%%%%%%%%%%%%%%%%%%%%%%%%%%%%%%%%%%%%%%%%%%%%%%%%%%%%%%%%%%%%%%%%%%%%%
% APPENDIX
%%%%%%%%%%%%%%%%%%%%%%%%%%%%%%%%%%%%%%%%%%%%%%%%%%%%%%%%%%%%%%%%%%%%%%%%%%%%%%%
%%%%%%%%%%%%%%%%%%%%%%%%%%%%%%%%%%%%%%%%%%%%%%%%%%%%%%%%%%%%%%%%%%%%%%%%%%%%%%%
\newpage
\appendix
\onecolumn

\begin{figure*}
    \subfigure[The Soft Heaviside function of Eq. \ref{eq:soft-heaviside}.]{
        \includegraphics[width=0.31\textwidth]{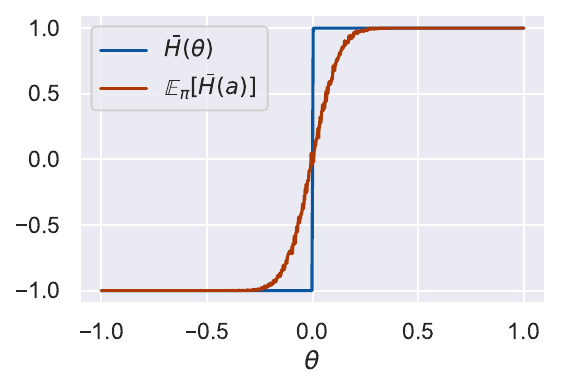}
        \label{fig:app-heaviside}
    }
    \hfill
    \subfigure[Gradient estimates at $N=1000$.]{
        \includegraphics[width=0.31\textwidth]{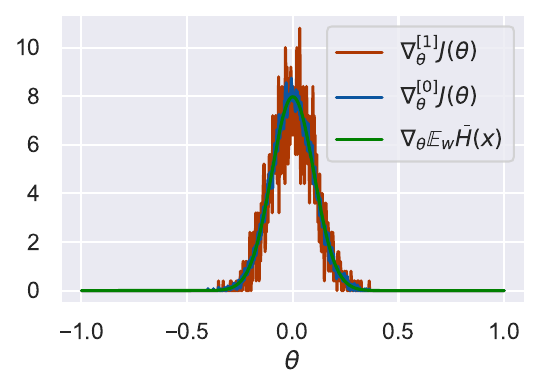}
        \label{fig:app-heaviside-grad}
    }
    \hfill
    \subfigure[Gradient sample error for different sample sizes $N$. Estimated by comparing the difference between the gradient estimate and the true gradient below.]{
        \includegraphics[width=0.31\textwidth]{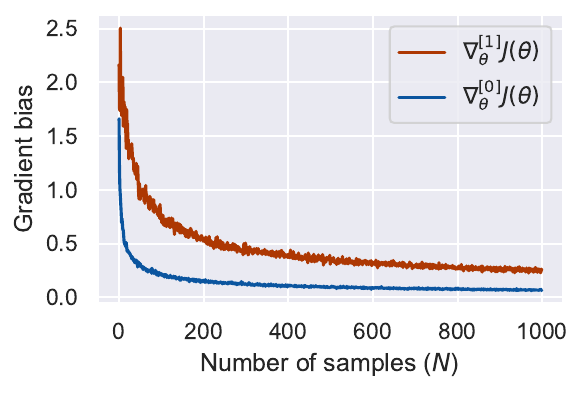}
        \label{fig:app-heaviside-bias}
    }
    \caption{Gradient sample error study for the Soft Heaviside function shown in Eq. \ref{eq:soft-heaviside}. Both ZOBG and FOBG exhibit sample errors at low sample sizes; however, FOBGs are especially susceptible to the "empirical bias" phenomena.}
    \label{fig:app_heaviside}
\end{figure*}

\section{Heaviside example} \label{app:heaviside}
This appendix provides additional details on the Heaviside example used to obtain intuition regarding FOBG sample error in Section \ref{sec:toy_example}.
\begin{align*}
    \bar{H}(x) =
    \begin{cases}
        1 & x > \nu/2 \\
        2x/\nu & |x| \leq \nu/2 \\
        -1 & x < -\nu/2
    \end{cases}
\end{align*}
Under stochastic input $x \sim \pi_\theta(\cdot) = \theta + w$ where $w \sim \mathcal{N}(0,\sigma)$, we can obtain the expected value:
\begin{align*}
    \E[w]{\bar{H}(x)} &= \int_{-\infty}^{\infty} \bar{H}(x) \pi_\theta(x) dx \\
    &= -\int_{-\infty}^{-\nu/2} \pi_\theta(x) dx + \int_{-\nu/2}^{\nu/2} \dfrac{2x}{\nu} \pi_\theta(x) dx + \int_{\nu/2}^\infty \pi_\theta(x) dx \\
    &= -\dfrac{1}{2} \erfc{\dfrac{\nu + 2\theta}{2 \sqrt{2} \nu}} +\dfrac{1}{2} \erfc{\dfrac{\nu - 2\theta}{2 \sqrt{2} \nu}} +\dfrac{\theta}{\nu} \erf{\dfrac{\nu - 2\theta}{2\sqrt{2}\sigma}} + \dfrac{\theta}{\nu} \erf{\dfrac{\nu+2\theta}{2\sqrt{2}\sigma}} \\
    & \quad \quad + \dfrac{\sigma \sqrt{2}}{\nu \sqrt{\pi}} \Big( \exp{-\dfrac{(\nu+2\theta)^2}{8\sigma^2}} - \exp{-\dfrac{(\nu-2\theta)^2}{8\sigma^2}} \Big)
\end{align*}
From the expectation, we can obtain the gradient w.r.t. the parameter of interest:
\begin{align*}
    \nabla_\theta \E[w]{\bar{H}(x)} &= \dfrac{1}{\sqrt{2\pi} \sigma} \exp{\Big(\dfrac{-(\nu+2\theta)^2}{8\sigma^2} \Big)} + \dfrac{1}{\sqrt{2\pi} \sigma} \exp{\Big( \dfrac{-(\nu-2\theta)^2}{8\sigma^2} \Big)} + \dfrac{1}{\nu} \erf{\dfrac{\nu-2\theta}{2\sqrt{2}\sigma}} + \dfrac{1}{\nu} \erf{\dfrac{\nu+2\theta}{2\sqrt{2}\sigma}} \\
    &\quad \quad -\dfrac{\sqrt{2} \theta}{\sqrt{\pi}\nu\sigma} \exp{\Big( \dfrac{-(\nu-2\theta)^2}{8 \sigma^2} \Big)} +\dfrac{\sqrt{2} \theta}{\sqrt{\pi}\nu\sigma} \exp{\Big( \dfrac{-(\nu+2\theta)^2}{8 \sigma^2} \Big)} \\
    &\quad \quad -\dfrac{1}{\sqrt{2}\sigma \nu} \exp{\big(-(\nu-2\theta)^{1/4\sigma^2} \big)} (\nu-2\theta)^{1/4\sigma^2 - 1} -\dfrac{1}{\sqrt{2}\sigma \nu} \exp{\big(-(\nu+2\theta)^{1/4\sigma^2}\big)} (\nu+2\theta)^{1/4\sigma^2 - 1}
\end{align*}

As seen from the equation above, the true gradient $\nabla_\theta \E[w]{\bar{H}(x)}  \neq 0$ at $\theta = 0$. However, using FOBG, we obtain $\nabla_\theta \bar{H}(a) = 0$ in samples where $|a| > \nu/2$, which occurs with probability at least $\nu/ \sigma \sqrt{2\pi}$. Even though both ZOBG and FOBG are theoretically unbiased as $N \rightarrow \infty$, both exhibit "empirical bias", as shown in Figure \ref{fig:app_heaviside}.

\newpage

\section{Proof of Lemma \ref{lem:bias-bound}}
\label{app:lemma-proof}

First we reiterate the assumptions to make this section self-sufficient and easier to read.

\begin{assumption} \label{ass:lipshitz-policy}
    Policy $\pi(\cdot | s) : \R^n \times \R^d \rightarrow [0,1]^m$ is continuously differentiable and Lipshitz smooth $\| \nabla_\vtheta \pi(a | s) \| \leq B_\pi$.
\end{assumption}

\begin{assumption} \label{ass:lipshitz-dynamics}
    Dynamics function $f(s,a) : \R^n \times \R^d \rightarrow \R^n$ is continuously differentiable and Lipshitz smooth in both arguments $\| \nabla_s f(s,a) \leq B_f$ and $\| \nabla_a f(s,a) \| \leq B_f$.
\end{assumption}

\begin{assumption} \label{ass:lipshitz-reward}
    Reward function $r(,a) : \R^n \times \R^d \rightarrow \R$ is continuously differentiable and Lipshitz smooth $\| \nabla_s r(s,a) \leq B_r$.
\end{assumption}

\begin{assumption} \label{ass:baseline}
    ZOBG use baseline $b$ subtraction which does not introduce gradient sample error  $\hat{\nabla}_\theta^[0](\theta) = \sum_{h=1}^H \big( r(s_h, a_h) - b \big) \nabla_\theta \log \pi_\theta (a_h | s_h)$ 
\end{assumption}

\begin{proof}
First, we expand our definition of sample error and define a random variable of a single Monte-Carlo sample
\begin{align} \label{eq:proof-pt1}
    B = \norm{\E{\bar{\nabla}_\vtheta^{[1]} J(\vtheta)} - \E{\bar{\nabla}_\vtheta^{[0]} J(\vtheta)}} &= \norm{\dfrac{1}{N} \sum_{i=1}^N \hat{\nabla}_\vtheta^{[1]} J_i(\vtheta) - \dfrac{1}{N} \sum_{i=1}^N\hat{\nabla}_\vtheta^{[0]} J_i(\vtheta) } \nonumber \\
    &= \dfrac{1}{N} \norm{\sum_{i=1}^N (\hat{\nabla}_\vtheta^{[1]} J_i(\vtheta) - \hat{\nabla}_\vtheta^{[0]} J_i(\vtheta))} \nonumber \\
    &\leq \| \hat{\nabla}^{[1]}_\vtheta J(\vtheta) - \hat{\nabla}^{[0]}_\vtheta J(\vtheta) \|
\end{align}

We drop the sample subscript $i$ for simplicity and assume that $\va \sim \pi_\vtheta (\cdot | \operatorname{sg}(\vs))$ where $\operatorname{sg}(\cdot)$ is the stop-gradient operator makes the expansion of $\hat{\nabla}^{[1]} J(\vtheta)$ easier \citep{deng2024facing}.

\begin{align} \label{eq:proof-pt2}
    &\hat{\nabla}^{[1]}_\vtheta J(\vtheta) - \hat{\nabla}_\vtheta^{[0]} J(\vtheta) \nonumber \\
    &= \nabla_\vtheta \sum_{h=1}^H r(\vs_h, \va_h) - \sum_{h=1}^H \big( r(\vs_h, \va_h) - b \big) \nabla_\vtheta \log \pi_\vtheta (\va_h | \vs_h) \nonumber \\
    &= \sum_{h=1}^H \nabla_{\va_h} r(\vs_h, \va_h) \nabla_\vtheta \pi (\va_h | \vs_h) + \bigg( \sum_{h'=1}^{h-1} \big( \prod_{t=h'+1}^{h-1} \nabla_{\vs_t} f(\vs_t, \va_t) \big) \nabla_{a_{h'}} f(\vs_{h'}, \va_{h'})^T \nabla_\vtheta \pi_\vtheta (\va_{h'} | \vs_{h'}) \bigg)^T \nabla_{\vs_h} r(\vs_h, \va_h) \nonumber \\ & \qquad - \sum_{h=1}^H \big( r(\vs_h, \va_h) - b \big) \nabla_\vtheta \log \pi_\vtheta (\va_h | \vs_h) \nonumber \\
    &= \sum_{h=1}^H \nabla_\vtheta \pi_\vtheta (\va_h | \vs_h)^T \bigg( \nabla_{\va_h} r(\vs_h, \va_h) - \big( r(\vs_h, \va_h) - b \big) \pi_\vtheta (\va_h | \vs_h)^{\circ - 1} \bigg) + \nonumber \\ & \qquad \bigg( \sum_{h'=1}^{h-1} \big( \prod_{t=h'+1}^{h-1} \nabla_{\vs_t} f(\vs_t, \va_t) \big) \nabla_{\va_{h'}} f(\vs_{h'}, \va_{h'}) \nabla_\vtheta \pi (\va_{h'} | \vs_{h'}) \bigg)^T \nabla_{\vs_h} r(\vs_h, \va_h)
\end{align}

where $\textbf{x}^{\circ-1}$ is the Hadamard inverse. Setting $b=r(\vs_h, 0)$, we can exploit the Lipshitz smoothness of $r$. In general, for any function $f(\vx)$:
\begin{align} \label{eq:smoothness-prop}
    f(\vy) &\leq f(\vx) + \nabla f(\vx)^T(\vy - \vx) + \dfrac{L}{2} \| \vy - \vx \|^2 \nonumber \\
    f(0) &\leq f(\vx) - \nabla f(\vx)^T \vx + \dfrac{L}{2} \| \vx \|^2 \nonumber \\
    f(\vx) - f(0) &\geq \nabla f(\vx)^T \vx - \dfrac{L}{2} \| \vx \|^2 \nonumber \\
    f(\vx) - f(0) &\geq (\nabla f(\vx)^T - \dfrac{L}{2} \vx^T) \vx
\end{align}

Applying $b=r(\vs_h, 0)$ to Eq. \ref{eq:proof-pt2} yields

\begin{align} \label{eq:proof-pt3}
    &= \sum_{h=1}^H \nabla_\vtheta \pi_\vtheta (\va_h | \vs_h)^T \bigg( \nabla_{\va_h} r(\vs_h, \va_h) - \big( r(\vs_h, \va_h) - r(\vs_h, 0) \big) \pi_\vtheta (\va_h | \vs_h)^{\circ - 1} \bigg) + \nonumber \\ & \qquad \bigg( \sum_{h'=1}^{h-1} \big( \prod_{t=h'+1}^{h-1} \nabla_{\vs_t} f(\vs_t, \va_t) \big) \nabla_{\va_{h'}} f(\vs_{h'}, \va_{h'}) \nabla_\vtheta \pi (\va_{h'} | \vs_{h'}) \bigg)^T \nabla_{\vs_h} r(\vs_h, \va_h) \nonumber \\
    &= \sum_{h=1}^H \nabla_\vtheta \pi_\vtheta (\va_h | \vs_h)^T \big( \nabla_{\va_h} r(\vs_h, \va_h) - \nabla_{\va_h} r(\vs_h, \va_h) + \dfrac{B_r}{2} \pi_\vtheta (\va_h | \vs_h) \big) \nonumber \\ & \qquad+ \bigg( \sum_{h'=1}^{h-1} \big( \prod_{t=h'+1}^{h-1} \nabla_{\vs_t} f(\vs_t, \va_t) \big) \nabla_{\va_{h'}} f(\vs_{h'}, \va_{h'}) \nabla_\vtheta \pi (\va_{h'} | \vs_{h'}) \bigg)^T \nabla_{\vs_h} r(\vs_h, \va_h) \nonumber \\
    &= \dfrac{B_r}{2} \sum_{h=1}^H \nabla_\vtheta \pi_\vtheta (\va_h | \vs_h)^T \pi_\vtheta (\va_h | \vs_h) + \bigg( \sum_{h'=1}^{h-1} \big( \prod_{t=h'+1}^{h-1} \nabla_{\vs_t} f(\vs_t, \va_t) \big) \nabla_{\va_{h'}} f(\vs_{h'}, \va_{h'}) \nabla_\vtheta \pi (\va_{h'} | \vs_{h'}) \bigg)^T \nabla_{\vs_h} r(\vs_h, \va_h)
\end{align}

Plug Eq. \ref{eq:proof-pt3} into Eq. \ref{eq:proof-pt1}:

\begin{align*}
    B &\leq \norm{\hat{\nabla}_\vtheta^{[1]} J(\vtheta) - \hat{\nabla}_\vtheta^{[0]} J(\vtheta)} \\
    &= \norm{\dfrac{B_r}{2} \sum_{h=1}^H \nabla_\vtheta \pi_\vtheta (\va_h | \vs_h)^T \pi_\vtheta(\va_h | \vs_h) + \bigg( \sum_{h'=1}^{h-1} \prod_{t=h'+1}^{h-1} \nabla_{\vs_t} f(\vs_t, \va_t) \nabla_{\va_{h'}} f(\vs_{h'}, \va_{h'}) \nabla_\vtheta \pi (\va_{h'} | \vs_{h'}) \bigg)^T \nabla_{\vs_h} r(\vs_h, \va_h)} \\
    & \qquad \qquad\text{Apply Eq. \ref{eq:smoothness-prop}} \\
    &\leq \norm{ \dfrac{B_r}{2} \sum_{h=1}^H \nabla_\vtheta \pi_\vtheta (\va_h | \vs_h)^T \pi_\vtheta(\va_h | \vs_h) } \\ & \qquad \qquad + \norm{ \bigg( \sum_{h'=1}^{h-1} \big( \prod_{t=h'+1}^{h-1} \nabla_{\vs_t} f(\vs_t, \va_t) \big) \nabla_{\va_{h'}} f(\vs_{h'}, \va_{h'}) \nabla_\vtheta \pi (\va_{h'} | \vs_{h'}) \bigg)^T \nabla_{\vs_h} r(\vs_h, \va_h ) } \\
    & \qquad \qquad \text{since $ \norm{ \nabla_\vtheta \pi_\vtheta(\va_h | \vs_h) } \leq B_\pi$ and $\norm{ \pi_\vtheta (\va_h | \vs_h) } \leq 1$} \\
    &\leq \dfrac{1}{2} H B_r B_\pi + \norm{ \bigg( \sum_{h'=1}^{h-1} \big( \prod_{t=h'+1}^{h-1} \nabla_{\vs_t} f(\vs_t, \va_t) \big) \nabla_{\va_{h'}} f(\vs_{h'}, \va_{h'}) \nabla_\vtheta \pi (\va_{h'} | \vs_{h'}) \bigg)^T \nabla_{\vs_h} r(\vs_h, \va_h) } \\
    &\leq \dfrac{1}{2} H B_r B_\pi + (H-1) B_r B_\pi B_f^{H-1} \\ 
    &\leq \dfrac{1}{2} H B_r B_\pi + H B_r B_\pi B_f^{H-1} \\
    &= H B_r B_\pi (\dfrac{1}{2} + B_f^{H-1})
\end{align*}
\end{proof}

\newpage

\section{Summary of modifications} \label{app:differences}

To develop the Adaptive Horizon Actor-Critic (AHAC) algorithm, we used the Short Horizon Actor-Critic (SHAC) algorithm \citep{xu2022accelerated} as a starting point. This section details all modifications applied to the SHAC in order to derive AHAC and achieve the reported results in this paper. We also note that some of these are not exclusive to either approach approach.

\begin{enumerate}
    \item \textbf{Adaptive horizon objective} - instead of optimizing the short horizon rollout return, we introduce the new constrained objective shown in Equation \ref{eq:actor-update-constrained}. To optimise that and adapt the horizon $H$, we introduced the dual problem in Equation \ref{eq:lagrangian} and optimised it directly for policy parameters $\vtheta$ and the Lagrangian coefficients $\vphi$.
    \begin{align*}
        \underbrace{\underbrace{J(\vtheta) := \sum_{h=t}^{t+T-1} \gamma^{h-t} r(\vs_h, \va_h) + \gamma^t V_\vpsi(\vs_{t+T})}_{\text{SHAC objective}} \quad s.t. \quad \| \nabla f(\vs_t, \va_t) \| \leq C \quad \forall t \in \{0, .., T\}}_{\text{AHAC objective}}
    \end{align*}
    \item \textbf{Double critic} - the original implementation of SHAC struggled with more complex tasks such as Humanoid due to its highly non-convex value landscape. The authors of \citep{xu2022accelerated} solved that by introducing a delayed target critic similar to prior work in deep RL \citep{lillicrap2015continuous}. We found that approach brittle and required more hyper-parameter tuning. Instead, we replaced it with a double critic similar to SAC \citep{haarnoja2018soft}. For our work, we found that it reduced the variance of asymptotic rewards achieved by AHAC while removing a hyperparameter. While this technique is usually applied to off-policy algorithms, we find it helpful in highly parallelized simulations due to the high data throughput.
    \item \textbf{Critic training until convergence} - empirically we found that different problems present different value landscapes. The more complex the landscape, the more training the critic required and the critic often failed to fit the data with the limited number of critic iterations done in SHAC (16). Instead of training the critic for a fixed number of iterations, we trained the (dual) critic of AHAC until convergence defined by $\sum_{i=n-5}^n \mathcal{L}_i(\vpsi) - \mathcal{L}_{i-1}(\vpsi) < 0.5$ where $\mathcal{L}_i(\vpsi)$ is the critic loss for mini-batch iteration $i$. We allowed the critic to be trained for a maximum of 64 iterations. We found that this resulted in asymptotic performance improvements on more complex tasks such as Humanoid and SNU Humanoid, while removing yet another hyper-parameter.
\end{enumerate}

\section{AHAC-1 algorithm} \label{app:ahac-1}

Algorithm \ref{alg:ahac-1} shows the single-environment version of AHAC that was described in Section \ref{sec:ahac-1}. While this algorithm applies the contact truncation technique perfectly and avoids all stiff contact, it is also not vectorizable. When attempting to vectorize AHAC-1, it necessitates cutting off compute graphs per-environment based on the individual environment dynamics. This is impossible to accomplish with typical deep learning frameworks such as PyTorch. Another alternative would be to execute different environments in different threads, but unfortunately, that does not benefit from GPU acceleration. 

\begin{algorithm}[h]
   \caption{Adaptive Horizon Actor-Critic 1 (Single environment)}
   \label{alg:ahac-1}
\begin{algorithmic}[1]
    \State \textbf{Given}: $\gamma$: discount rate
    \State \textbf{Given}: $\alpha_\vtheta, \alpha_\vpsi$: learning rates
    \State \textbf{Given}: $H$: maximum trajectory length
    \State \textbf{Given}: $C$: contact threshold
    \State \textbf{Initialize learnable parameters} $\vtheta, \vpsi$
    \State $t \leftarrow 0$
    \vskip 6pt
    
    \While{episode not done}
        \State \(\triangleright\) Rollout policy
        \State \text{Initialize rollout buffer } $D$ 
        \While{ $\| \nabla f \| \leq C \text{ and } h \leq H$}
            \State $\va_t \sim \pi_\vtheta(\cdot | \vs_t)$
            \State $r_t = r(\vs_t, \va_t)$
            \State $s_{t+1} = f(\vs_t, \va_t)$
            \State $D \leftarrow D \cup \{(\vs_{t+h}, \va_{t+h}, r_{t+h}, V_\vpsi(\vs_{t+h+1}))\}$
            \State $t \leftarrow t + 1$
        \EndWhile
        \vskip 6pt

        \State \(\triangleright\) Train actor using Eq. \ref{eq:actor-update}
        \State $\vtheta \leftarrow \vtheta - \alpha_\vtheta \nabla_\vtheta J(\vtheta) $ 
        \vskip 6pt

        \State \(\triangleright\)Train critic using Eq. \ref{eq:critic-update} 
        \While{not converged} 
            \State sample $(\vs, \hat{V}(\vs)) \sim D$
            \State $\vpsi \leftarrow \vpsi + \alpha_\vpsi \nabla_\vpsi \mathcal{L}(\vpsi)$
        \EndWhile
    \EndWhile
\end{algorithmic}
\end{algorithm}

\section{Differentiable Simulation Setup: dflex} \label{app:simulation}
The experimental simulator, \textit{dflex} \citep{xu2022accelerated}, employed in Section \ref{sec:experiments}, is a GPU-based differentiable simulator utilizing the Featherstone formulation for forward dynamics and a spring-damper contact model with Coulomb friction.

The dynamics function $f$ is modeled by solving the forward dynamics equations:
\begin{align*}
    M \Ddot{q} = J^T \mathcal{F} (q, \dot{q}) + c(q, \dot{q}) + \tau (q, \dot{q}, a)
\end{align*}
where, $q, \dot{q}, \Ddot{q}$ are joint coordinates, velocities, and accelerations, respectively. $\mathcal{F}$ represents external forces, $c$ includes Coriolis forces, and $\tau$ denotes joint-space actuation. Mass matrix $M$ and Jacobian $J$ are computed concurrently using one thread per-environment. The composite rigid body algorithm (CRBA) is employed for articulation dynamics, enabling caching of the matrix factorization for reuse in the backward pass through parallel Cholesky decomposition.

After determining joint accelerations $\ddot{q}$, a semi-implicit Euler integration step updates the system state $s = (q, \dot{q})$. Torque-based control is employed for simple environments, where the policy outputs $\tau$ at each timestep. For further details, see \citep{xu2022accelerated}. It is noted that \textit{dflex} is no longer actively developed and has been succeeded by \textit{warp} \citep{warp2022}.

\newpage
\section{Environment details} \label{app:env-details}

\begin{figure}
     \centering
     \begin{subfigure}
         \centering
         \includegraphics[width=0.19\textwidth]{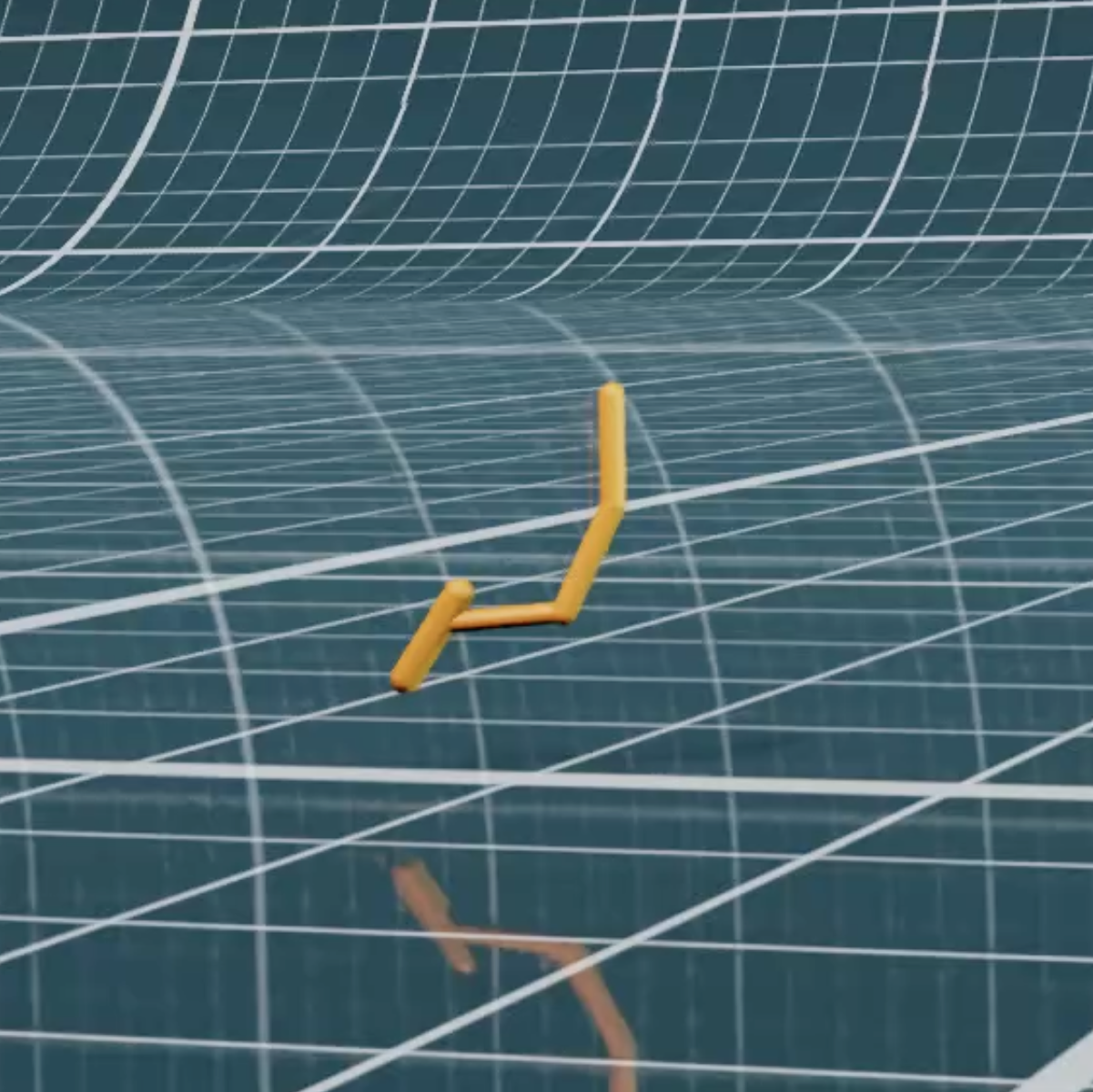}
     \end{subfigure}
     \hfill
     \begin{subfigure}
         \centering
         \includegraphics[width=0.19\textwidth]{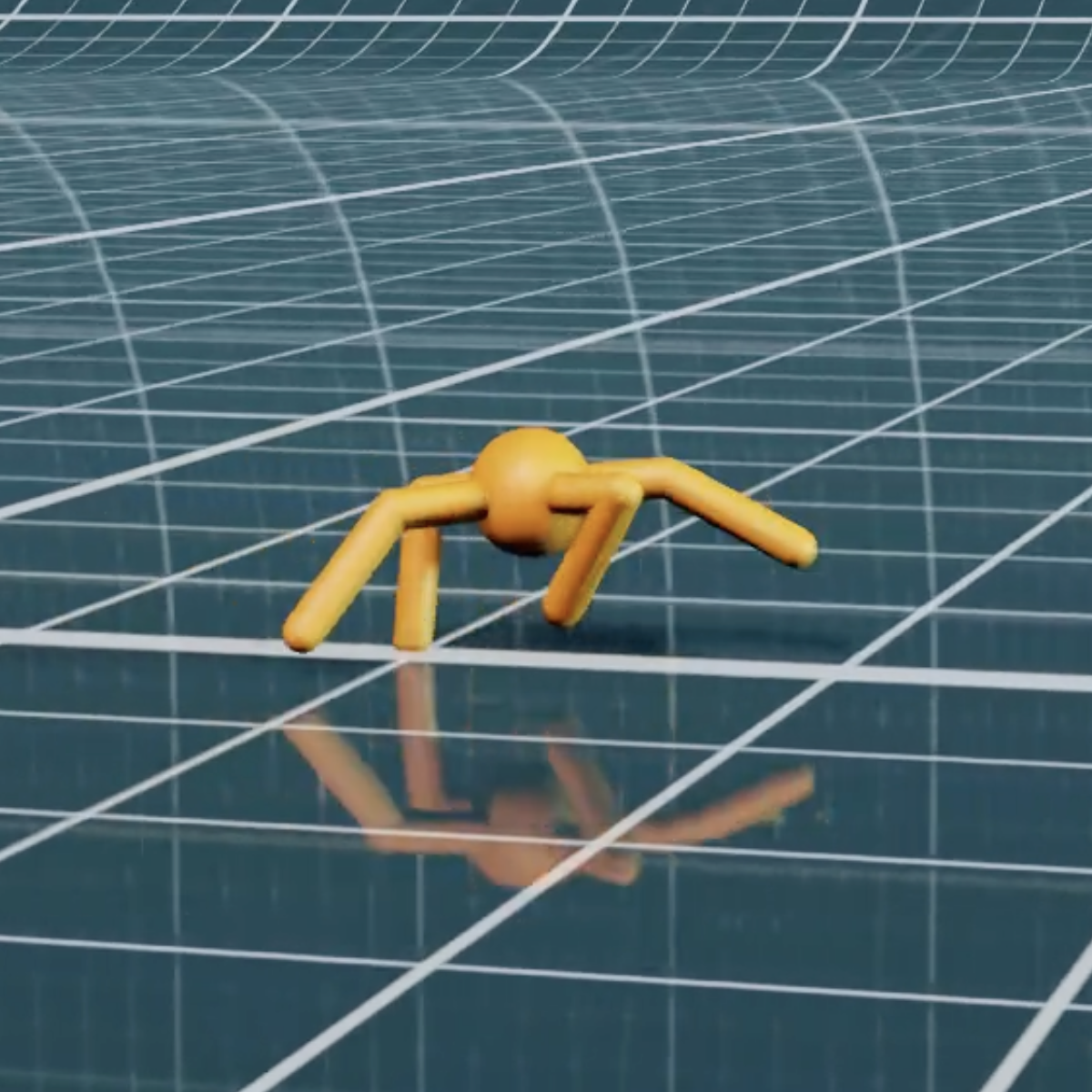}
     \end{subfigure}
     \hfill
     \begin{subfigure}
         \centering
         \includegraphics[width=0.19\textwidth]{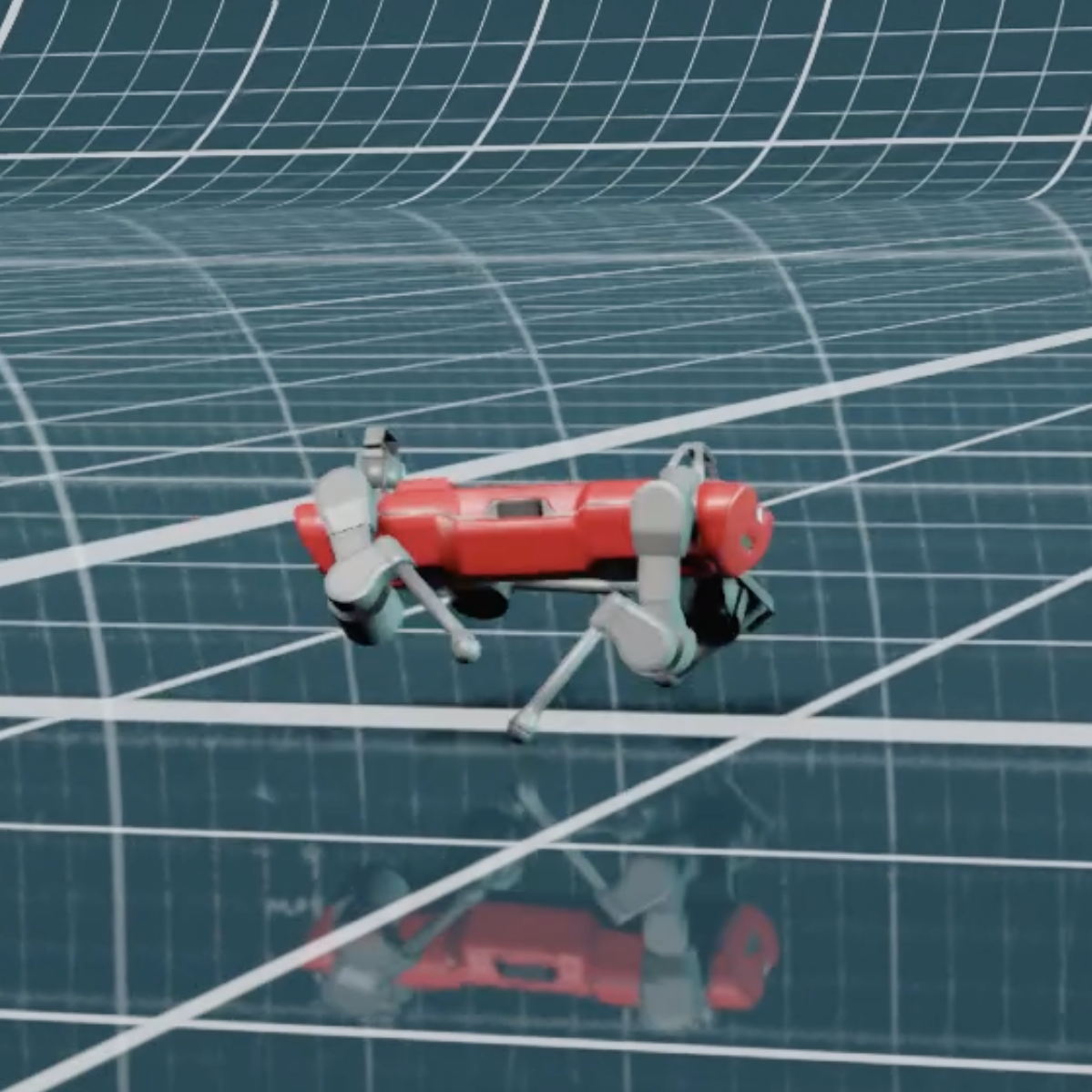}
     \end{subfigure}
     \hfill
     \begin{subfigure}
         \centering
         \includegraphics[width=.19\textwidth]{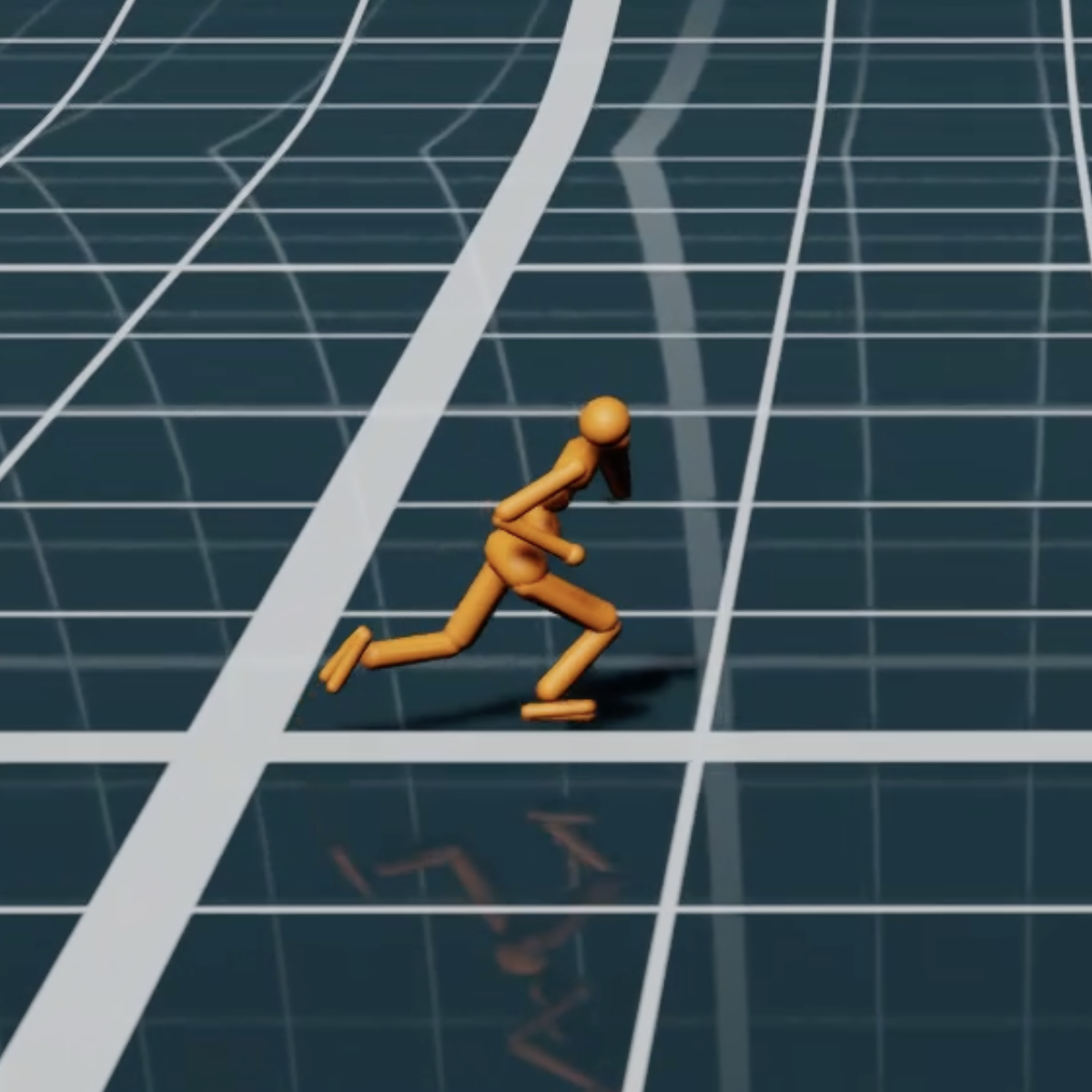}
     \end{subfigure}
         \hfill
     \begin{subfigure}
         \centering
         \includegraphics[width=0.19\textwidth]{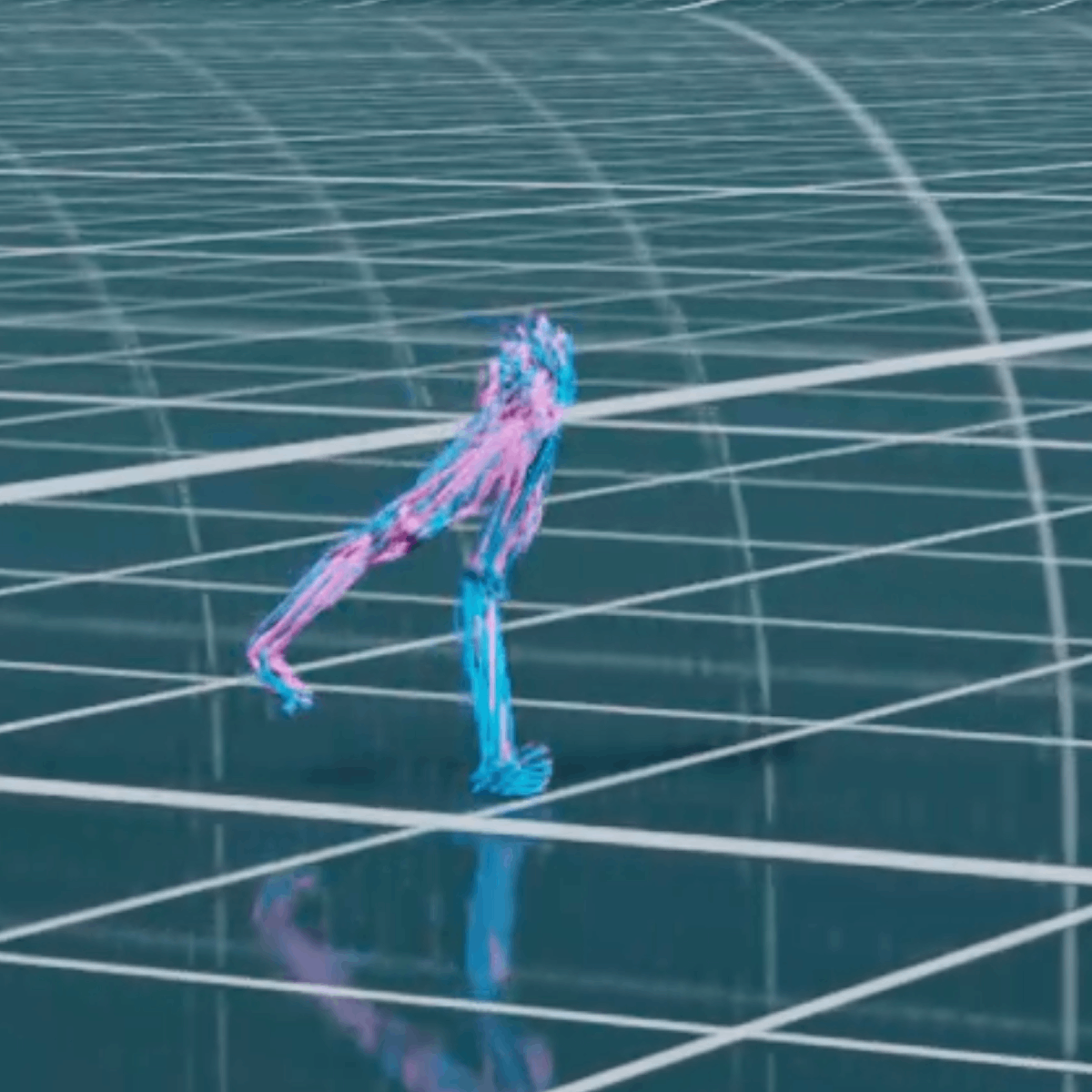}
     \end{subfigure}
        \caption{Locomotion environments (left to right): Hopper, Ant, Anymal, Humanoid and SNU Humanoid.}
        \label{fig:envs_viz}
\end{figure}

In this paper, we explore 5 locomotion tasks with increasing complexity. They are described below and shown in Figure \ref{fig:envs_viz}.

\begin{enumerate}
    \item \textbf{Hopper}, a single-legged robot jumping only in one axis with $n=11$ and $m=3$.
    \item \textbf{Ant}, a four-legged robot with $n=37$ and $m=8$.
    \item \textbf{Anymal}, a more sophisticated quadruped with $n=49$ and $m=12$ modeled after \citep{hutter2016anymal}. 
    \item \textbf{Humanoid}, a classic contact-rich environment with $n=76$ and $m=21$, which requires extensive exploration to find a good policy.
    \item \textbf{SNU Humanoid}, a version of Humanoid lower body where instead of joint torque control, the robot is controlled via $m=152$ muscles intended to challenge the scaling capabilities of algorithms.
\end{enumerate}

All tasks share the same common main objective - maximize forward velocity $v_x$:

\begin{table}[h]
\centering
\begin{NiceTabular}{ll}
\CodeBefore
  \rowcolor[HTML]{D6D6EB}{1}
  \rowcolors[HTML]{2}{eaeaf3}{FFFFFF} % seaborn colors
  % \rowcolors{2}{gray!15}{white}
\Body
\toprule
Environment & Reward    \\ \midrule
Hopper & $v_x + R_{height} + R_{angle} - 0.1\| \va \|^2_2$ \\
Ant & $v_x + R_{height} + 0.1 R_{angle} + R_{heading} - 0.01\| \va \|^2_2$ \\
Anymal & $v_x + R_{height} + 0.1 R_{angle} + R_{heading} - 0.01\| \va \|^2_2$ \\
Humanoid & $v_x + R_{height} + 0.1 R_{angle} + R_{heading} - 0.002\| \va \|^2_2$ \\
Humanoid STU & $v_x + R_{height} + 0.1 R_{angle} + R_{heading} - 0.002\| \va \|^2_2$ \\
\bottomrule
\end{NiceTabular}
\caption{Rewards used for each task benchmarked in Section \ref{sec:experiments}}
\label{tab:rewards}
\end{table}

We additionally use auxiliary rewards $R_{height}$ to incentivize the agent to, $R_{angle}$ to keep the agent's normal vector point up, $R_{heading}$ to keep the agent's heading pointing towards the direction of running and a norm over the actions to incentivize energy-efficient policies. For most algorithms, none of these rewards, apart from the last one, are crucial to succeeding in the task. However, all of them aid learning policies faster.

\begin{align*}
    R_{height} = \begin{cases}
        h - h_{term} & if h \geq h_{term} \\
        -200 (h - h_{term})^2 & if h < h_{term} 
    \end{cases}
\end{align*}

\begin{align*}
    R_{angle} = 1 - \bigg( \dfrac{\theta}{\theta_{term}} \bigg)^2
\end{align*}

$R_{angle} = \| \vq_{forward} - \vq_{agent} \|_2^2$ is the difference between the heading of the agent $\vq_{agent}$ and the forward vector $\vq_{agent}$. $h$ is the height of the CoM of the agent and $\theta$ is the angle of its normal vector. $h_{term}$ and $\theta_{term}$ are parameters that we set for each environment depending on the robot morphology. Similar to other high-performance RL applications in simulation, we find it crucial to terminate episode early if the agent exceeds these termination parameters. However, it is worth noting that AHAC is still capable of solving all tasks described in the paper without these termination conditions, albeit slower.

\section{Hyper-parameters} \label{app:hyperparams}

This section details all hyper-parameters used in the main experiments in Section \ref{sec:experiments}. PPO and SAC, as our MFRL baselines, have been tuned to perform well across all tasks, including task-specific hyper-parameters. SVG has not been specifically tuned for all benchmarks due to time limitations but instead uses the hyper-parameters presented in \citep{amos2021model}.\footnote{Tuning SVG proved difficult as we were unable to vectorize the algorithm, resulting in up to 2-week training times. This made it difficult to tune for our benchmarks} SHAC is tuned to perform well across all tasks using a fixed $H=32$ as in the original work \citep{xu2022accelerated}. AHAC shares all of its common hyper-parameters with SHAC and only has its horizon learning rate $\alpha_\vphi$ tuned per-task. The contact threshold $C$ and iterative critic training criteria did not benefit from tuning. Note that the double critic employed by AHAC uses the same hyper-parameters used by the SHAC critic. Therefore, we have left AHAC under-tuned in comparison to SHAC in order to highlight the benefits of the adaptive horizon mechanism presented in this work. 

Table \ref{tab:common-hyperparamters} shows common hyper-parameters shared between all tasks. While table \ref{tab:task-hyperparams} shows hyper-parameters specific to each problem,. Where possible, we attempted to use the hyper-parameters suggested by the original works; however, we also attempted to share hyper-parameters between algorithms to ease comparison. If a specific hyperparameter is not mentioned, then it is the one used in the original work behind the specific algorithm.

\begin{table}[h]
\centering
\begin{NiceTabular}{llllll}
\CodeBefore
  \rowcolor[HTML]{D6D6EB}{1}
  \rowcolors[HTML]{2}{eaeaf3}{FFFFFF} % seaborn colors
  % \rowcolors{2}{gray!15}{white}
\Body
\toprule
                         & AHAC & SHAC & PPO  & SAC    & SVG    \\ \midrule
Mini-epochs              &      & 16   & 5    &        & 4      \\ 
Batch size               & 8    & 8    & 8    & 32     & 1024   \\
$\lambda$   & 0.95 & 0.95 & 0.95 &        &        \\ 
$\gamma$    & 0.99 & 0.99 & 0.99 & 0.99   & 0.99   \\ 
H - horizon                       &      & 32   & 32   &        & 3      \\ 
C - contact thresh.                       &  500 &    &    &        &       \\ 
Grad norm                & 1.0  & 1.0  & 1.0  &        &        \\ 
$\epsilon$  &      &      & 0.2  &        &        \\ 
Actor $log(\sigma)$ bounds           &      &      &      & (-5,2) & (-5,2) \\
$\alpha$ - temperature   &      &      &      & 0.2    & 0.1    \\ 
$\lambda_\alpha$ &      &      &      & $10^{-4}$   & $10^{-4}$   \\
$|D|$ - buffer size                    &      &      &      & $10^6$    & $10^6$    \\ 
Seed steps               & 0    & 0    & 0    & $10^4$  & $10^4$ \\  \bottomrule
\end{NiceTabular}
\caption{Table of hyper-parameters for all algorithms benchmarked in Section \ref{sec:experiments}. These are shared across all tasks.}
\label{tab:common-hyperparamters}
\end{table}

\begin{table}[h]
\centering
\begin{NiceTabular}{llllll}
\CodeBefore
  \rowcolor[HTML]{D6D6EB}{1}
  \rowcolors[HTML]{2}{eaeaf3}{FFFFFF} % seaborn colors
\Body
\toprule
              & Hopper        & Ant           & Anymal     & Humanoid   & SNU Humanoid \\ \midrule
Actor layers  & (128, 64, 32) & (128, 64, 32) & (256, 128) & (256, 128) & (512, 256)   \\ 
Actor $\alpha_\vtheta$     & $2 \times 10^{-3}$          & $2 \times 10^{-3}$          & $2 \times 10^{-3}$       & $2 \times 10^{-3}$       & $2 \times 10^{-3}$         \\
Horizon $\alpha_\vphi$     & $2 \times 10^{-4}$          & $1 \times 10^{-5}$          & $1 \times 10^{-5}$       & $1 \times 10^{-5}$       & $1 \times 10^{-5}$         \\ 
Critic layers & (64, 64)      & (64, 64)      & (256, 128) & (256, 128) & (256, 256)   \\ 
Critic $\alpha_\vpsi$     & $4 \times 10^{-3}$          & $2 \times 10^{-3}$          & $2 \times 10^{-3}$       & $5 \times 10^{e-4}$       & $5 \times 10^{-4}$         \\
Critic $\tau$    & 0.2           & 0.2           & 0.2        & 0.995      & 0.995        \\ \bottomrule
\end{NiceTabular}
\caption{Task-specific hyper-parameters. All benchmarked algorithms share the same actor and critic network hyper-parameters with ELU activation functions. AHAC and PPO do not have target critic networks and, as such, do not have $\tau$ as a hyper-parameter.}
\label{tab:task-hyperparams}
\end{table}

\newpage
\section{Experimental results} \label{app:tabular-results}

\begin{figure}[h]
  \centering
  \includegraphics[width=\linewidth]{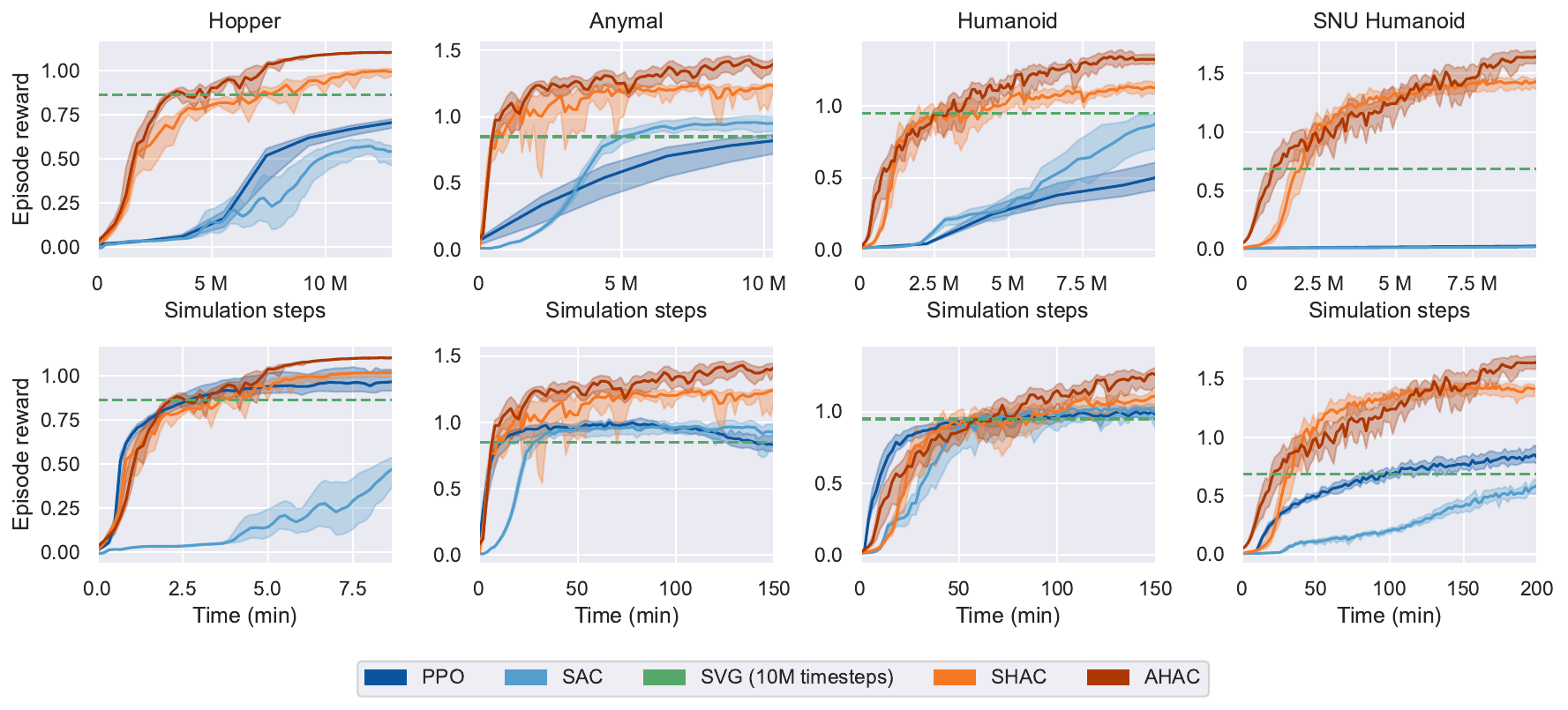}
  \caption{\textbf{Reward curves for all tasks against both simulation steps and training time}. We plot 50 \% IQM and 95 \% CI.}
  \label{fig:experiments}
\end{figure}

In addition to the experimental results in Section \ref{sec:experiments}, here we provide the same results in more detail. Figure \ref{fig:experiments} depicts step-wise and time-wise reward curves for all experiments. Tables \ref{tab:results} and \ref{tab:results_raw} provide asymptotic (converged) results for all tasks with PPO-normalized and raw rewards, respectively.

\begin{table}[h]
\centering
\begin{NiceTabular}{cccccc}
\CodeBefore
  \rowcolor[HTML]{D6D6EB}{1}
  \rowcolors[HTML]{2}{eaeaf3}{FFFFFF} % seaborn colors
\Body
\toprule
     & Hopper      & Ant         & Anymal      & Humanoid    & SNU Humanoid \\ \midrule
PPO  & $1.00 \pm 0.11$ & $1.00 \pm 0.12$ & $1.00 \pm 0.03$ & $1.00 \pm 0.05$ & $1.00 \pm 0.09$   \\ 
SAC  & $0.87 \pm 0.16$ & $0.95 \pm 0.08$ & $0.98 \pm 0.06$ & $1.04 \pm 0.04$ & $0.88 \pm 0.11$  \\ \midrule
SVG  & $0.84 \pm 0.08$ & $0.83 \pm 0.13$ & $0.84 \pm 0.19$ & $1.06 \pm 0.16$ & $0.75 \pm 0.23$  \\ \midrule
SHAC & $1.02 \pm 0.03$ & $1.16 \pm 0.13$ & $1.26 \pm 0.04$ & $1.15 \pm 0.04$ & $1.44 \pm 0.08$ \\ 
AHAC & $1.10 \pm  0.00$ & $1.41 \pm 0.08$ & $1.46 \pm 0.06$ & $1.35 \pm 0.07$ & $1.64 \pm 0.07$  \\ \bottomrule
\end{NiceTabular}
\caption{Tabular results of the asymptotic rewards achieved by each algorithm across all tasks. The results presented are PPO-normalized 50 \% IQM and standard deviation across 10 random seeds. All algorithms have been trained until convergence.}
\label{tab:results}
\end{table}

% perf_scales
% {'ant': 6604.663357204862,
%  'hopper': 4742.296956380208,
%  'anymal': 12028.991672092014,
%  'humanoid': 7292.7856852213545,
%  'humanoidsnu': 4114.013700544121}

\begin{table}[h]
\centering
\begin{NiceTabular}{cccccc}
\CodeBefore
  \rowcolor[HTML]{D6D6EB}{1}
  \rowcolors[HTML]{2}{eaeaf3}{FFFFFF} % seaborn colors
\Body
\toprule
     & Hopper      & Ant         & Anymal      & Humanoid    & SNU Humanoid \\ \midrule
PPO  & $4742 \pm 521$ & $6605 \pm 793$ & $12029 \pm 360$ & $7293 \pm 365$ & $4114 \pm 370$   \\ 
SAC  & $4126 \pm 759$ & $6275 \pm 528$ & $11788 \pm 722$ & $7285 \pm 292$ & $3620 \pm 453$  \\ \midrule
SVG  & $3983 \pm 379$ & $5482 \pm 859$ & $10104 \pm 2286$ & $7731 \pm 1167$ & $3086 \pm 946$  \\ \midrule
SHAC & $4837 \pm 142$ & $7662 \pm 859$ & $15157 \pm 481$ & $8387 \pm 292$ & $5924 \pm 329$ \\ 
AHAC & $5216 \pm  21$ & $9313 \pm 528$ & $17562 \pm 722$ & $9846 \pm 511$ & $6746 \pm 288$  \\ \bottomrule
\end{NiceTabular}
\caption{Tabular results of the asymptotic (end of training) rewards achieved by each algorithm across all tasks. The results presented are 50 \% IQM and standard deviation across 10 random seeds. All algorithms have been trained until convergence.}
\label{tab:results_raw}
\end{table}

\newpage
\section{Ablation study details} \label{app:ablation-study}

In Section \ref{sec:experiments}, we provided an ablation study of the individual contributions of our proposed approach, AHAC, as summarized in Appendix \ref{app:differences}. In this section, we provide further details on the conducted experiments. The aim of the study is to understand what changes contribute to the asymptotic performance of AHAC. To best achieve that, we started with SHAC as the baseline, using the tuned version detailed in Appendix \ref{app:hyperparams} above. Afterwards, we add the individual components that contribute to AHAC using the hyper-parameter from the section above. Note that only hyper-parameters particular to AHAC have been tuned to achieve the results presented in this paper; all other hyper-parameters are the ones tuned to our baseline SHAC with $H=32$. In particular, we have only tuned the adaptive horizon learning rate $\alpha_\vpsi$ and contact threshold $C$. Table \ref{tab:ablation-hyperparams} shows the detailed differences between the ablations presented in Section \ref{sec:experiments}. The ablations include:

\begin{enumerate}[nosep]
    \item SHAC H=32 - our baseline with most hyper-parameters tuned to it.
    \item SHAC H=29 - SHAC using the horizon $H$ which AHAC converges to asymptotically.
    \item Adaptive Objective - SHAC using the adaptive horizon objective introduced in Eq. \ref{eq:lagrangian} but without using it to adapt to the horizon.
    \item Adaptive Horizon - SHAC using the objective in Eq \ref{eq:lagrangian} and adapting the horizon. This is equivalent to AHAC without the double critic and with iterative training.
    \item Iterative critic - SHAC with a single target critic, utilizing iterative critic training until convergence.
    \item Double critic - SHAC with a double critic and no target. 
\end{enumerate}

\begin{table}[h]
\centering
\begin{NiceTabular}{@{}llllll@{}}
\CodeBefore
  \rowcolor[HTML]{D6D6EB}{1}
  \rowcolors[HTML]{2}{eaeaf3}{FFFFFF} % seaborn colors
\Body
\toprule
                                                                            & Ablation         & H        & Actor objective & Critic           & \begin{tabular}[c]{@{}l@{}}Iterative \\critic training\end{tabular} \\ \midrule
                                                                    & SHAC H=32        & 32       & Eq. \ref{eq:actor-update}           & Single w/ target &                           \\ \midrule
\Block[fill=white!100]{3-1}{\begin{tabular}[c]{@{}l@{}}Actor\\ ablations\end{tabular}}  & SHAC H=29        & 29       & Eq. \ref{eq:actor-update}           & Single w/ target &                           \\
                                                                            & Adapt. Objective & 32       & Eq. \ref{eq:lagrangian}           & Single w/ target &                           \\
                                                                            & Adapt. Horizon   & adaptive & Eq. \ref{eq:lagrangian}           & Single w/ target &                           \\ \midrule
\Block[fill=white!100]{2-1}{\begin{tabular}[c]{@{}l@{}}Critic\\ ablations\end{tabular}} & Iterative critic & 32       & Eq. \ref{eq:actor-update}           & Single w/ target & \checkmark \\
                                                                            & Double critic    & 32       & Eq. \ref{eq:actor-update}           & Dual           &                           \\ \midrule
                                                                            & AHAC             & adaptive & Eq. \ref{eq:lagrangian}           & Dual           & \checkmark \\ \bottomrule
\end{NiceTabular}
\caption{Differences between ablations studied, split into actor and critic ablations. All ablations only introduce one component to the baseline, SHAC.}
\label{tab:ablation-hyperparams}
\end{table}

\begin{figure}[h]
    \centering
    \includegraphics[width=\linewidth]{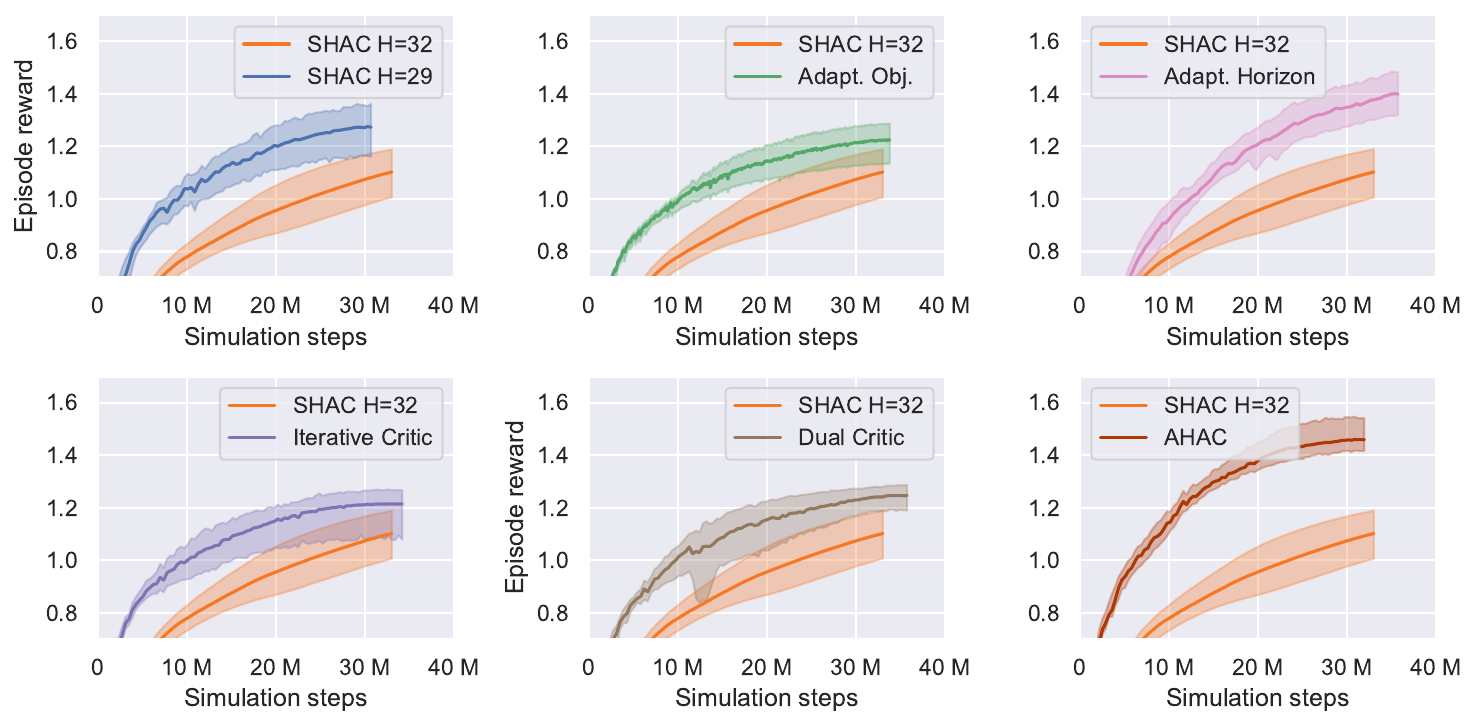}
    \caption{Standalone ablation results for the Ant task. These results are the same as in Figure \ref{fig:ablation} but presented in a different format for improved legibility.}
    \label{fig:ablations-standalone}
\end{figure}

Previously in Section \ref{sec:experiments}, we only provided end of training results for the Ant task. In Table \ref{tab:ablation-table} we provide the same results in tabular form. We also provide the learning curves for the same experiments in Figure \ref{fig:ablations-standalone}.

% \usepackage{booktabs}
\begin{table}[htb]
\centering
\begin{NiceTabular}{@{}ll@{}}
\CodeBefore
  \rowcolor[HTML]{D6D6EB}{1}
  \rowcolors[HTML]{2}{eaeaf3}{FFFFFF} % seaborn colors
\Body
\toprule
Ablation           & Asymptotic reward \\ \midrule
SHAC H=32          & 1.16 $\pm$ 0.14      \\
1. SHAC H=29          & 1.23 $\pm$ 0.17      \\
2. Adaptive Objective & 1.18 $\pm$ 0.18      \\
3. Adaptive Horizon   & 1.35 $\pm$ 0.12      \\
4. Iterative Critic   & 1.17 $\pm$ 0.13      \\
5. Double Critic        & 1.20 $\pm$ 0.07      \\
AHAC               & 1.41 $\pm$ 0.08      \\ \bottomrule
\end{NiceTabular}
\caption{Results of asymptotic performance of our ablation study showing 50\% IQM and standard deviation.}
\label{tab:ablation-table}
\end{table}

\end{document}